\documentclass[10pt,twocolumn,letterpaper]{article}

\usepackage[pagenumbers]{paper}

\usepackage{amsthm, mathtools, physics}
\usepackage[final,nopatch=footnote]{microtype}
\usepackage{siunitx}
\usepackage{setspace, wrapfig, nicefrac, adjustbox, multirow, pifont, footmisc}
\usepackage[scr=boondox]{mathalpha}

\AtEndPreamble{\crefname{appendix}{Sec.}{Secs.}}   %
\AtEndPreamble{\crefname{theorem}{Thm.}{Thms.}}

\makeatletter
\@ifpackageloaded{lineno}{%
  \IfFormatAtLeastTF{2025-06-01}{%
    \RemoveFromHook{build/column/before}[lineno]%
    \AtBeginDocument{%
      \def\@LN@makecol{%
        \@LN@orig@makecol
        \setbox\@outputbox\vbox{%
          \boxmaxdepth\@maxdepth
          \protected@write\@auxout{}{\string\def\string\@LN@column{\if@firstcolumn1\else2\fi}}%
          \box\@outputbox}}}%
  }{}%
}{}
\makeatother

\definecolor{linkblue}{rgb}{0.21,0.49,0.74}
\usepackage[pagebackref,breaklinks,colorlinks,allcolors=linkblue,bookmarks=false]{hyperref}
\newcommand{\cosd}[1][d]{\mathcal{C}_{#1}} %

\newcommand{\aed}[1]{\begin{aligned} #1 \end{aligned}}
\newcommand{\beq}[1]{\begin{equation}#1\end{equation}}
\newcommand{\beqs}[1]{\begin{equation*}#1\end{equation*}}

\newcommand{\inner}[2]{\left\langle #1,#2 \right\rangle}
\newcommand{\rbr}[1]{\left(#1\right)}
\newcommand{\sbr}[1]{\left[#1\right]}

\newcommand{\abr}[1]{\left|#1\right|}
\providecommand{\norm}{} %
\renewcommand{\norm}[1]{\left\|#1\right\|}

\providecommand\f[2]{\ensuremath \frac{#1}{#2}}

\let\P\relax
\DeclareMathOperator{\P}{\mathbb{P}}
\DeclareMathOperator*{\E}{\mathbb{E}}

\DeclareMathOperator*{\cov}{\text{Cov}}

\newtheorem{theorem}{Theorem}
\newtheorem{proposition}{Proposition}
\newtheorem{lemma}{Lemma}

\theoremstyle{definition}   %
\newtheorem{remark}{Remark}
\theoremstyle{plain}

\def \d {\delta}
\def \s {\sigma}

\def \r {\rho}

\def \th {\theta}
\def \a {\alpha}
\def \b {\beta}

\def \e {\varepsilon}

\def \l {\lambda}

\def \i {\dot{\iota}}
\def\bb{\mathscr{b}}
\def\vv{\mathscr{v}}
\def\kk{\mathscr{k}}

\def \D {\Delta}

\def \Om {\Omega}

\def \reals {\mathbb{R}}

\graphicspath{{images/}{./}}

\title{LeJEPA Features Are Not Isotropic, and Should Not Be}

\author{Richeek Das, Pratik Chaudhari\\
GRASP Laboratory, University of Pennsylvania\\
{\tt\small \{richeek, pratikac\}@seas.upenn.edu}
}

\newlength{\mylen}
\makeatletter
\renewcommand\paragraph{\@startsection{paragraph}{4}{\z@}{1.25ex \@plus 0.5ex \@minus 0.2ex}{-1em}{\normalfont\normalsize\bfseries}}
\DeclareMathSizes{9.95}{9}{7}{5}
\DeclareMathSizes{10}{9}{7}{5}
\DeclareMathSizes{11}{10}{8}{6}
\makeatother

\begin{document}

\maketitle
{\renewcommand{\thefootnote}{}\footnotetext{Project page: \url{https://www.engineering.upenn.edu/~richeek/polarreg/}}}

\begin{abstract}

The SIGReg term in LeJEPA regularizes features to have an isotropic Gaussian distribution. Our first result is to show that this regularizer, in practice, cannot achieve its goal. We design a modified regularizer to enforce the objective better. This leads to slightly improved performance. For example, on ImageNet-100, our modified regularizer reaches the final accuracy of SIGReg in 1.3$\times$ fewer epochs.
Our second result argues that self-supervised encoders should not have isotropic Gaussian features. We show that the alignment term in LeJEPA conflicts with SIGReg, because it couples the magnitude of the features with their directions. In contrast, isotropic Gaussian features would entail independent magnitude and directions. Features obtained with LeJEPA are, in fact, not isotropic.
So LeJEPA has two issues. Its regularizer is weak, and the isotropic Gaussian target is too strict.
Our third result is about reformulating the isotropic Gaussian regularizer to enforce only the marginal distributions of the magnitude and direction, without enforcing independence. This leads to a simple regularization term, PolarReg, that can be optimized reliably across many different datasets, batch sizes and embedding dimensions without any hyper-parameter changes.
On ImageNet-1k, our technique matches the accuracy of LeJEPA and its variants in 4.7$\times$ fewer epochs. It improves the final accuracy by 1.8 points.

\end{abstract}

\section{Introduction}
\label{sec:intro}

\begin{figure}[t]
    \centering
    \vspace{-0.5\baselineskip}
    \includegraphics[width=\linewidth]{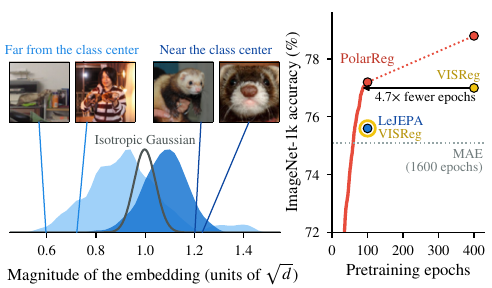}
    \caption{
    \textbf{Left:}
    LeJEPA embeddings do not have an isotropic Gaussian distribution.
    The distribution of embedding magnitudes for black-footed ferret images, from a ViT-B encoder trained using LeJEPA on ImageNet-1k.
    Histograms show this distribution for 25\% of images nearest (dark) to and farthest (light) from the class center in cosine similarity.
    Far away images tend to be ambiguous and have embeddings with smaller magnitudes.
    \textbf{Right:} We develop a new regularizer, named PolarReg, that replaces SIGReg in LeJEPA.
    On ImageNet-1k, a ViT-L encoder trained with PolarReg reaches the final accuracy of LeJEPA and its variants like VISReg in 86 epochs, 4.7$\times$ fewer.
    }
    \label{fig:teaser}
\end{figure}

Joint-embedding predictive architectures (JEPA)~\cite{lecun2022path} are trained to predict the embedding of one view, such as a randomly cropped image, from the embedding of another view of the same image.
Feature collapse, i.e., features that can be predicted reliably across different views but are not sufficiently informative of the downstream task, is a major issue while instantiating this idea. Popular methods like DINO~\cite{caron2021dino}, iBOT~\cite{zhou2022ibot}, and I-JEPA~\cite{assran2023ijepa} use heuristics to work around this issue.
More recently, LeJEPA~\cite{lejepa2025} and its variants like VISReg~\cite{wu2026visreg} argue that they are effective at preventing feature collapse, without these heuristics, because of a specific regularization term.

The LeJEPA objective consists of an alignment term and a regularization term. The alignment term pulls together the embeddings of different views of the same image. The regularization term, called sketched isotropic Gaussian regularization (SIGReg), is computed over a batch of images. The original authors argue that it is a test statistic for whether the embeddings are independent samples from a standard Gaussian distribution.
More formally, let $f_\th: X \to \reals^d$ denote the encoder that maps inputs $x \in X$ to embeddings $u \in \reals^d$ using weights $\th$.
For a batch of $\bb$ images $x_1, \dots, x_{\bb}$, each with $\vv$ augmented views $x_i^v$, let $u_i^v = f_\th(x_i^v)$ be the embedding of view $v$ of image $i$.
The LeJEPA objective
\beq{
    \aed{
        \ell(\th) &= \frac{1}{\bb \vv d} \sum_{i=1}^\bb \sum_{v=1}^\vv \norm{u_i^v - m_i}^2 + \Om_{\text{SIGReg}}, \text{ where}\\
        \Om_{\text{SIGReg}} &= \frac{\l \bb}{\vv \kk} \sum_{v=1}^\vv \sum_{k=1}^\kk \int w(t) \abr{\varphi_k^v(t) - e^{-t^2/2}}^2 \dd{t}.
    }
    \label{eq:lejepa}
}
The first term is the alignment term, where $m_i = {\vv}^{-1} \sum_v u_i^v$ is the mean embedding of the views of image $i$.
The second term, SIGReg, encourages the embeddings to have an isotropic Gaussian distribution.
It projects the embeddings of a view onto $\kk$ random unit vectors $a_1, \dots, a_\kk \in \reals^d$ called slices and $\varphi_k^v(t) = \bb^{-1} \sum_i \operatorname{exp}(\i t a_k^\top u_i^v)$ is the empirical characteristic function of these projections.
The integral corresponds to the Epps--Pulley test statistic~\cite{epps1983test} against the characteristic function of the standard Gaussian $\text{N}(0,1)$, weighted by a function $w(t)$ to keep the integral finite.
By the Cram\'er--Wold theorem, as $\kk \to \infty$, matching such projections to $\text{N}(0,1)$ is equivalent to matching the distribution of embeddings to $\text{N}(0, I_{d\times d})$.
\textbf{Our first result is to show that SIGReg only weakly enforces its objective in practice (see \cref{rem:app:pr}). SIGReg is small even when the embedding distribution is quite different from an isotropic Gaussian. We prove that using more slices does not fix this (\cref{sec:blindness}), even though the regularizer is sound as $\kk \to \infty$.}

The authors of LeJEPA argue~\cite[Theorem 1]{lejepa2025} that embeddings should be distributed as an isotropic Gaussian because the downstream task is not known a priori. Indeed, if the embeddings are anisotropic, their projections onto different unit vectors have different variances and a linear probe for the task can be more biased.
However, this claim holds only in the worst case over all downstream tasks. Natural tasks have structure and can benefit from anisotropic embeddings~\cite{ramesh2024many}. \textbf{Our second result is to show that SIGReg conflicts with the alignment term in LeJEPA.}
An isotropic Gaussian embedding distribution makes the magnitudes and directions of the embeddings independent of each other. We show that the alignment term instead introduces dependence between the two. In fact, the minimizer of the alignment term assigns smaller magnitudes to more ambiguous images (see \cref{fig:teaser,sec:dependence}). \textbf{We therefore argue that, in LeJEPA-like objectives, the magnitudes of the embeddings should depend on their directions.} We show that this enables the encoder to train faster and more accurately. We prove that such embeddings do not collapse (\cref{rem:app:coupling}).

Since the two terms in LeJEPA work against each other, the accuracy of the approach depends heavily on the regularization weight $\l$.
On ImageNet-100, the classification accuracy of the original LeJEPA encoders varies by up to 7.7\% points for different values of $\l$. The best $\l$ depends on the number of views $\vv$, the number of slices $\kk$ and how the integral is approximated~\cite[Table 7]{lejepa2025}.
On ImageNet-1k, the accuracy varies by up to 3.8\% points depending upon these settings~\cite[Table 1]{lejepa2025}.
LeJEPA uses hyper-parameter search to choose these settings. A new dataset, architecture, number of views or embedding dimension may need a new search.

\textbf{Our third result is to develop a new regularizer, called PolarReg, that replaces SIGReg.}
This regularizer matches the marginal distributions of the embedding magnitudes, and the pairwise inner products between embedding directions, to those of an isotropic Gaussian, computed analytically.
PolarReg is very simple; it is designed with optimization techniques to ensure that its hyper-parameters need no tuning for different batch sizes, embedding dimensions or magnitudes (\cref{sec:autodose}). \textbf{All experiments in this paper use a single hyper-parameter setting for PolarReg.}
On ImageNet-1k, PolarReg matches the accuracy of VISReg in 4.7$\times$ fewer epochs, and reaches a final accuracy 1.8\% better than VISReg (\cref{fig:teaser,sec:res:accuracy}). It uses the same training hyper-parameters as VISReg, with no additional search.
Finally, we show that the magnitude of embeddings obtained with our approach, serves as a confidence score and reliably detects out-of-domain images without any labels (\cref{sec:res:radius}).
Related work is discussed in \cref{app:related}.

\section{Methods}
\label{sec:methods}

\subsection{SIGReg cannot detect collapsed embeddings}
\label{sec:blindness}

\begin{figure}[t]
    \centering
    \vspace{-0.75\baselineskip}
    \includegraphics[width=\linewidth]{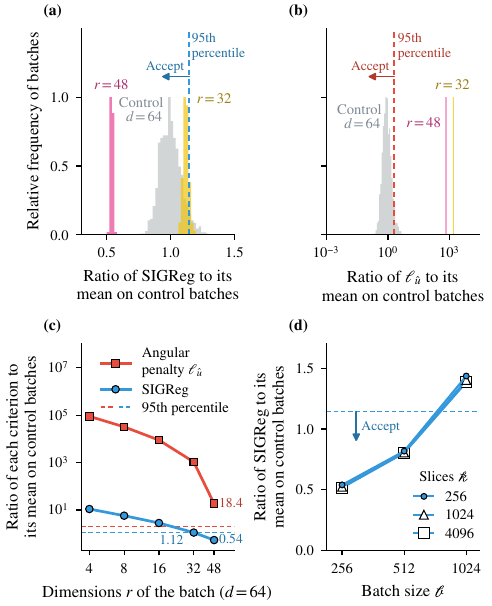}
    \caption{\textbf{SIGReg cannot sufficiently distinguish between samples of an isotropic Gaussian distribution and rank-collapsed embeddings that minimize SIGReg.}
    The criteria are shown as ratios relative to their mean values on the control batches.
    \textbf{(a)} SIGReg computed on 1000 control batches of $\bb = 256$ samples of $\text{N}(0, I_{64 \times 64})$ (gray) and on batches restricted to $r = 48$ (pink) and $r = 32$ (yellow) dimensions. The Epps--Pulley statistic in SIGReg accepts all the pink batches and 90\% of the yellow batches.
    \textbf{(b)} The angular penalty $\ell_{\hat u}$ in \cref{eq:angular} on the same batches as in (a). It correctly rejects the rank-collapsed embeddings.
    \textbf{(c)} SIGReg and the angular penalty $\ell_{\hat u}$, each minimized over batches in an $r$-dimensional subspace.
    The dashed lines indicate the acceptance thresholds at the 95th percentile of the respective quantity.
    The angular penalty rejects low-rank embeddings. Both criteria grow as $r$ decreases.
    \textbf{(d)} SIGReg, as in (c), for $r = 48$, computed for different numbers of slices $\kk$ against the batch size $\bb$.
    Only with a large enough batch size $\bb = 1024$, SIGReg correctly rejects the rank-collapsed embeddings.
    }
    \label{fig:rank_blindness}
    \vspace{-0.75\baselineskip}
\end{figure}

Consider \cref{eq:lejepa} with the number of views $\vv = 1$. SIGReg is a test statistic for the null hypothesis that $u_1, \dots, u_\bb$ are independent samples of $\text{N}(0, I_{d \times d})$.
Suppose a batch of embeddings consists of $u_i = B z_i$ where $B \in \reals^{d \times r}$ is a uniformly random matrix with orthonormal columns and $z_i \in \reals^r$ for $i = 1, \dots, \bb$. This batch has a rank of at most $r$. If we minimize SIGReg using gradient descent on $z_i$ initialized from $z_i \sim \text{N}(0, I_{r \times r})$, we can find batches of rank at most $r$ that are deemed maximally similar to an isotropic Gaussian distribution by SIGReg. Embeddings that have an isotropic Gaussian distribution by construction act as controls for this experiment. If the Epps--Pulley statistic in SIGReg accepts such batches at a level say 0.05, this would show that SIGReg can lead to collapsed embeddings, not distributed as an isotropic Gaussian.

\Cref{fig:rank_blindness}a,c show that SIGReg deems embeddings with rank as little as 75\% of the embedding dimension as having an isotropic Gaussian distribution. In fact, the values of SIGReg for such batches are lower than on any of the control batches, e.g., pink histogram in \cref{fig:rank_blindness}a with rank $r=48$ for $d = 64$. Even when the embeddings span only a 32-dimensional subspace out of $d=64$ dimensions, SIGReg accepts 90\% of such batches as being isotropic. \Cref{fig:rank_blindness}d shows that using more slices $\kk$ does not make the test more discriminative, but using a larger batch size $\bb$ does.

\Cref{prop:sliced_power} explains this observation.
We consider embeddings sampled from $\text{N}(0, \Sigma)$ where $\Sigma \in \reals^{d \times d}$ has $r$ nonzero eigenvalues, all equal to $d/r$.
We show that the expected SIGReg on such embeddings exceeds its expectation on control batches by a term that grows with the batch size $\bb$.
For intuition, consider projections of the $\bb$ embeddings along a particular slice.
The variance of projections along a slice $a$ is $a^\top \Sigma a$. We show that the variance of this quantity across random slices $a$ equals $2(d - r)/(r(d + 2))$. The empirical variance of $\bb$ samples of $\text{N}(0, 1)$ has variance about $2/\bb$.
Therefore, SIGReg can identify non-isotropic embeddings if $2(d - r)/(r(d + 2))$ and $2/\bb$ are significantly different from each other. These quantities do not depend on the number of slices $\kk$, which explains why the discriminative ability of SIGReg does not improve with a larger $\kk$. This also shows why the two quantities can differ a lot for large batch sizes $\bb$.
For a fixed ratio $r/d$, the first variance falls as $1/d$.
Therefore, SIGReg is less discriminative for higher-dimensional embeddings. For example, at $d = 256$, the test accepts a batch of embeddings where 75\% of the dimensions are collapsed.

This failure of SIGReg also manifests in LeJEPA training.
As \cref{fig:teaser} shows, LeJEPA embeddings are anisotropic.
In \cref{rem:app:pr}, we measure this anisotropy using a quantity called the participation ratio, which is the effective dimensionality of the embeddings. It is $d$ only for isotropic embeddings. \Cref{fig:app:pr} shows that throughout training, the participation ratio of LeJEPA embeddings is much smaller than $d$.

\subsection{A polar decomposition of the isotropic Gaussian regularization}
\label{sec:decomposition}

The magnitude and direction of embeddings with an isotropic Gaussian distribution are independent of each other. In this section, we calculate analytical expressions for the marginal distributions of the magnitude and direction of such embeddings. This will enable us to enforce the isotropic Gaussian regularization more precisely.

\begin{theorem}[Polar decomposition of an isotropic Gaussian]
\label{thm:polar}
For $u \in \reals^d$, let $u = \rho \hat u$ where $\rho = \norm{u}$ is the magnitude and $\hat u = u/\norm{u}$ is a unit vector. A nonzero random vector $u \sim \text{N}(0, I_{d \times d})$ if and only if (i) $\rho$ is distributed according to the $\chi_d$ distribution, (ii) $\hat u$ is uniform on the $(d-1)$-dimensional sphere $S^{d-1}$, and (iii) $\rho$ and $\hat u$ are independent.
\end{theorem}
\noindent
\Cref{app:polar} provides the proof.
\Cref{thm:polar} shows that, for embeddings $u$ to be samples of an isotropic Gaussian, the marginal distribution of the magnitude must be $\chi_d$, while their directions should be distributed uniformly on the $(d-1)$-dimensional sphere.
By the third condition, magnitudes and directions are independent random variables, that is, every direction has the same marginal of the magnitudes, and vice versa. For intuition, note that for large $d$, the distribution $\chi_d$ concentrates around $\sqrt d$, that is the embeddings lie on a thin shell. Two points sampled independently and uniformly on a $(d-1)$-dimensional sphere are nearly orthogonal for large $d$. Their inner product has mean zero and variance $1/d$.

Using \cref{thm:polar}, we can set the regularization term in \cref{eq:lejepa} to be
\beq{
    \Om = \underbrace{\l_\r \ell_\r(\th) + \l_{\hat u} \ell_{\hat u}(\th)}_{
    \text{PolarReg-naive}} + \underbrace{\l_\perp \ell_\perp(\th)}_{\text{independence term}}.
    \label{eq:full_decomposition}
}
where $\ell_\r$ and $\ell_{\hat u}$ enforce the conditions on the marginals of the magnitude and the direction, respectively, and $\ell_\perp$ enforces their independence.
The coefficients $\l_\r$, $\l_{\hat u}$ and $\l_\perp$ are hyper-parameters that we will discuss later. We will refer to \cref{eq:full_decomposition} with $\l_\perp = 0$, as $\Om_{\text{PolarReg-naive}}$.

The marginals in \cref{thm:polar} can be matched using any reasonable notion of divergence between two distributions. Since these constraints are enforced during training using empirical distributions of the embeddings, they need to scale well---both statistically and computationally---with the batch size $\bb$ and dimensionality $d$. We next discuss how to impose these constraints using only scalar quantities.

\paragraph{The magnitude regularizer $\ell_\r$.}
For two probability distributions supported on a one-dimensional space, the 2-Wasserstein distance $W_2$ is the $\ell^2$ distance between the inverses of their cumulative distribution functions (CDFs)~\cite[Theorem 2.18]{villani2003topics}.
It is zero if and only if the two distributions are equal~\cite[Theorem 7.3]{villani2003topics}.
Suppose we have $n$ samples $z = (z_1, \dots, z_n)$ with each scalar $z_i \sim P$. We can compute a differentiable approximation of $W_2^2(P, Q)$ given by,
\beq{
    W_2^2(z, Q) = \frac{1}{n} \sum_{r=1}^n \rbr{z_{(r)} - q_r}^2,
    \label{eq:w2}
}
where $z_{(1)} \leq \dots \leq z_{(n)}$ is the sorted sequence and $q_r = F_Q^{-1}((r - 1/2)/n)$ is the $r$-th of $n$ midpoint quantiles of the target distribution with CDF $F_Q$.
It is therefore straightforward to enforce $\rho \sim \chi_d$ in \cref{thm:polar}, using the 2-Wasserstein distance.
We apply \cref{eq:w2} to the magnitudes $\rho_i^v = \norm{u_i^v}$ of the embeddings in the batch, for all $\bb$ images and $\vv$ views, with target $\chi_d$,
\beq{
    \ell_\r(\th) = \frac{1}{\bb \vv} \sum_{r=1}^{\bb \vv} \rbr{\rho_{(r)} - q_r}^2,
    \label{eq:radial}
}
where $\rho_{(r)}$ is the $r$-th smallest magnitude and $q_r$ is the $r$-th quantile of $\chi_d$.

\paragraph{The angular regularizer $\ell_{\hat u}$.}
We show in \cref{thm:app:cosine} that directions $\hat u$ are uniformly distributed on the sphere if and only if the inner product of two independent directions follows a distribution $\cosd$ with density proportional to $(1 - s^2)^{(d-3)/2}$ on $[-1, 1]$. Equivalently, $(1 + s)/2 \sim \text{Beta}(\b, \b)$ with $\b = (d-1)/2$. 
The quantiles of $\cosd$ are therefore those of $\text{Beta}(\b, \b)$, rescaled to $[-1, 1]$.
We can enforce the embedding directions to be uniform on the sphere by matching the pairwise inner products $\cosd$.
Since different views of an image are not independent, we compute this regularizer on the directions $\hat m_i$ of the mean embeddings $m_i$ over the $\vv$ views of each image $i$.
Let $s_{ij} = \inner{\hat m_i}{\hat m_j}$ be the inner product between the directions of the mean embeddings of images $i$ and $j$, for $i < j$. 
We set the angular regularizer to the squared 2-Wasserstein distance \cref{eq:w2} between these inner products and the target distribution $\cosd$:
\beq{
    \ell_{\hat u}(\th) = \frac{2}{\bb(\bb-1)} \sum_{r=1}^{\bb(\bb-1)/2} \rbr{s_{(r)} - q_r}^2,
    \label{eq:angular}
}
where $s_{(r)}$ is the $r$-th smallest inner product and $q_r$ here is the $r$-th quantile of $\cosd$.

\paragraph{The independence regularizer $\ell_\perp$.}
Since magnitudes $\r$ and directions $\hat u$ have different dimensions, we enforce independence using the maximum mean discrepancy (MMD) between their joint distribution and the product of their marginals~\cite{JMLR:v13:gretton12a,gretton2005hsic}.
Let $\kappa_\r(\r, \r') = \exp(-(\r - \r')^2/(2\s_\r^2))$ and $\kappa_{\hat u}(\hat u, \hat u') = \exp(-\norm{\hat u - \hat u'}^2/(2\s_{\hat u}^2))$ be Gaussian kernels, and let $\kappa((\r, \hat u), (\r', \hat u')) = \kappa_\r(\r, \r')\, \kappa_{\hat u}(\hat u, \hat u')$.
Let $P$ be the joint distribution of $(\r, \hat u)$ and $P_\r P_{\hat u}$ the product of its marginals.
The squared MMD between them is
\beq{
    \text{MMD}^2 = \E\sbr{\kappa(z, z')} + \E\sbr{\kappa(w, w')} - 2\E \sbr{\kappa(z, w)},
    \label{eq:mmd}
}
where $z, z' \sim P$ and $w, w' \sim P_\rho P_{\hat u}$ are independent.
$\text{MMD}^2$ is zero if and only if $P = P_\rho P_{\hat u}$,
i.e., if and only if $\r$ and $\hat u$ are independent~\cite[Section 2.3]{song2012feature}.
We compute the expectations in \cref{eq:mmd} separately for each view $v$, over distinct images $i$ and $j$ in the batch. Let $K^v_{ij} = \kappa_\r(\r_i^v, \r_j^v)$ and $L^v_{ij} = \kappa_{\hat u}(\hat u_i^v, \hat u_j^v)$ for $i \neq j$.
We set
\beq{
\adjustbox{max width=0.9\linewidth}{$\displaystyle
\begin{aligned}[b]
    \ell_\perp(\th) = \frac{1}{\vv} \sum_{v=1}^\vv \Big[ & \frac{(\bb-2)!}{\bb!} \sum_{i, j} K^v_{ij} L^v_{ij} + \frac{(\bb-4)!}{\bb!} \sum_{i, j, p, q} K^v_{ij} L^v_{pq} \\
    & - \frac{2 (\bb-3)!}{\bb!} \sum_{i, j, p} K^v_{ij} L^v_{ip} \Big],
\end{aligned}$}
\label{eq:independence}
}
where each sum runs over pairwise distinct indices.
The bandwidths $\s_\r$ and $\s_{\hat u}$ are chosen as the median pairwise distances in the batch, and $\ell_\perp$ does not backpropagate through them or the magnitudes $\r$, see \cref{app:independence}.

\paragraph{Does this modified regularizer improve discriminative ability compared to SIGReg?}

\Cref{rem:app:second_moment} shows that on control batches in $d$ dimensions, the average squared inner product between pairs of directions is $1/d$ in expectation, with standard deviation $\sim 2/(d\bb)$.
If the embeddings span $r < d$ dimensions, it is at least $(\bb/r - 1)/(\bb - 1)$.
This bounds the angular regularizer from below, so even at small batch sizes, it stays far above its values on control batches, however the directions are arranged.
In \cref{fig:rank_blindness}b, the angular regularizer rejects every collapsed batch of embeddings that SIGReg accepts.
In \cref{fig:rank_blindness}c, the angular regularizer, minimized over lower-rank batches as in \cref{sec:blindness}, remains discriminative for all ranks $r$.

The angular term can easily detect duplicated images in a batch.
We show in \cref{rem:app:duplicates} that if a fraction $p$ of the embeddings are the same in a batch, SIGReg grows only by a factor $1 + 2p$ irrespective of the number of slices or batch size.
\Cref{app:duplicates} helps put this into perspective, if there is one repeated image in a batch of size $\bb = 256$ for $d = 64$ dimensional embeddings, SIGReg only increases 0.8\% while the angular regularizer increases 6.2 times.
This is important in practice because large image datasets often contain duplicates \cite{oquab2024dinov2}.

\begin{table}[t]
    \centering
    \caption{\textbf{SIGReg compared with two variants of \cref{eq:full_decomposition}, with and without the independence penalty.}
    Average linear probing accuracy for a ViT-B trained on ImageNet-100 for 100 epochs with $d = 256$, over 3 seeds.
    MMD$^2$ is reported here after dividing by its standard deviation over random shuffles of the magnitudes across images. This makes it comparable across methods (see \cref{app:independence}).
    }
    \label{tab:independence}
    \begin{adjustbox}{max width=\linewidth}
    \begin{tabular}{lrrr}
        \toprule
        Method & Accuracy (\%) & $\text{MMD}^2$ & Speedup \\
        \midrule
        SIGReg & 78.92 & 86.8 & 1$\times$ \\
        PolarReg-naive + Independence & 79.98 & 18.7 & 1.3$\times$ \\
        PolarReg-naive & 81.00 & 114.7 & 1.5$\times$ \\
        \bottomrule
    \end{tabular}
    \end{adjustbox}
\end{table}

\paragraph{The modified regularizer, PolarReg-naive, performs better than SIGReg.}

\Cref{tab:independence} compares SIGReg with two versions of \cref{eq:full_decomposition}; one with all three kinds of penalties (PolarReg-naive + Independence), and one without the independence penalty, $\l_\perp = 0$ (PolarReg-naive).
It reports the accuracy of linear probing on ImageNet-100, the independence measure between embedding magnitude and direction, and how much faster each reaches the final accuracy of SIGReg.
With all three penalties, accuracy is higher than with SIGReg, magnitudes and directions are more independent, and training is 1.3$\times$ faster.
Surprisingly, without the independence regularizer, accuracy improves further, and training is 1.5$\times$ faster than SIGReg. As expected, $\text{MMD}^2$ is much higher without the independence regularizer.
We will exploit this observation further. The following section argues that independence between magnitudes and directions is not necessary, and is in fact the wrong constraint to enforce.

\begin{figure}[t]
    \centering
    \includegraphics[width=\linewidth]{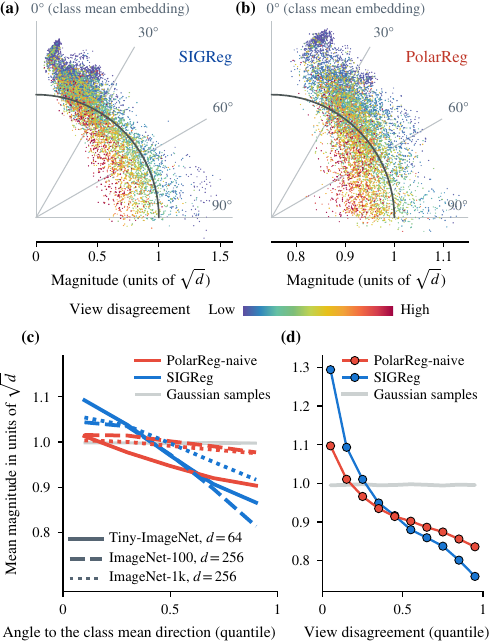}
    \caption{
    \textbf{Dependence of embedding magnitude and direction in trained encoders.}
    (a) and (b) are polar charts where, each point is an image from a six class subset of ImageNet-1k.
    A point lies at the magnitude of the image embedding and at its angle from the mean embedding of its class, which is at $0^\circ$.
    Color shows the view disagreement $D_i$ of each image, and the black arc marks magnitude $\sqrt d$.
    The embeddings are from ViT-B encoders trained on ImageNet-1k with $d = 256$.
    \textbf{(a)} Embeddings from SIGReg, same encoder as \cref{fig:teaser} (left). \textbf{(b)} Embeddings from PolarReg.
    \textbf{(c)} Within each class, embeddings are split in terms of quantiles of their relative angle to the mean embedding of the class. This plot shows the mean magnitude for each such quantile, averaged over classes.
    For both methods, the magnitude falls with increasing relative angle.
    \textbf{(d)} Mean magnitude of image embeddings for different quantiles of view disagreement. We use 10,000 validation images of Tiny-ImageNet-200 with $d = 64$ dimensional embeddings and $\vv = 8$ augmented views.
    }
    \label{fig:shape}
    \vspace{-0.5\baselineskip}
\end{figure}

\subsection{Embedding magnitudes and directions should not be independent of each other}
\label{sec:dependence}

We show in \cref{prop:rearrangement} that if all views of an image $i$ have the same magnitude $\r_i$, the alignment term in LeJEPA is proportional to $\sum_i \r_i^2 D_i$. Here $D_i = \sum_v \norm{\hat u_i^v - \bar u_i}^2$ is the view disagreement of image $i$ and $\hat u_i^v = u_i^v/\rho_i$ are the directions of the views and $\bar u_i$ is their mean over $\vv$ views.
The view disagreement $D_i$ varies with the embedding direction. It is large for ambiguous images that typically lie in between two semantically distinct classes, and it is small for embeddings that lie near their class centroid.
Since swapping the magnitudes of two images does not alter the distribution of the magnitudes, the alignment term is minimized only when images with larger view disagreement get smaller magnitude embeddings. This, in effect, makes the magnitude and direction dependent on each other.
Conceptually, this dependence encodes the structure in the data learned during self-supervised learning. Enforcing the embeddings to have an isotropic Gaussian distribution prevents this structure from crystallizing during training.

\Cref{fig:shape}a,b show that in encoders trained with SIGReg and PolarReg, the magnitude of an embedding falls as its angle to the mean embedding of its class grows. Additionally, images with larger view disagreement lie at larger angles, and at a given angle, they have smaller magnitudes.
\Cref{fig:app:weight} shows that images with larger view disagreement consequently get less weight in the alignment term.
\Cref{fig:shape}c and d measure these trends over whole datasets. In (c), averaged over classes, the magnitude falls as the angle to the class mean grows, for both SIGReg and PolarReg-naive on three datasets. In (d), on Tiny-ImageNet-200, it falls as the view disagreement grows.
Just like \cref{fig:teaser}, these results further corroborate our findings in \cref{sec:blindness}. SIGReg embeddings are anisotropic, their magnitudes depend strongly on the directions.
\Cref{app:rearrangement} discusses \cref{prop:rearrangement} and verifies its assumption on trained encoders.

\textbf{We therefore keep only the magnitude and angular regularizers and drop the independence term.}
We prove in \cref{rem:app:coupling} that forgoing the independence term cannot result in collapsed embeddings.
To elaborate, if the embedding magnitudes are distributed as $\chi_d$ and directions are uniform on the sphere, we show that every eigenvalue of the second-moment matrix of the embeddings is within $2/\sqrt d$ of 1, i.e., the embeddings do not collapse, whatever the dependence between magnitude and direction.

\subsection{Making PolarReg-naive independent of the batch size, embedding dimension and scale}
\label{sec:autodose}

The optimal coefficients $\l_\r$ and $\l_{\hat u}$ depend on the batch size $\bb$, embedding dimension $d$, and the structure in the data.
For example, when the embeddings are isotropic Gaussian, the angular penalty converges to zero as $1/\bb^2$~\cite[Theorems 5.1 and 5.9]{bobkov2019one}. Thus, a coefficient tuned on large batch sizes is stronger than necessary on small ones (\cref{app:autodose}).
It is possible to tune the coefficients once and scale them heuristically to be effective across different datasets. For example, we tuned $\l_{\hat u} = 20$ and $\l_\r = 0.045$ on Tiny-ImageNet-200 with $\bb = 256$ and $d = 64$ (see the ablation in \cref{tab:app:weights}).
\Cref{app:autodose} shows that for a new setting, we can rescale both coefficients so that the gradient norms of the alignment term and the two regularizers, with respect to the embeddings, keep their ratios from our original tuned setting.

We can develop the key ideas of PolarReg-naive to obtain a more robust approach.
\Cref{eq:radial,eq:angular} are minimized when the $r$-th smallest magnitude or inner product equals the $r$-th quantile of $\chi_d$ or $\cosd$, respectively.
So at each iteration, we can compute targets $u_i^{v*}$ for embeddings $u_i^v$ and move the latter towards the former.
Since both penalties are quadratic in the magnitudes and inner products, doing so is easy.
We discuss this idea next, see \cref{app:ot_construction} for more details.

\paragraph{Moving the embedding magnitudes.}
A batch has $\bb\vv$ embeddings, one for each image view. Suppose $\r_i^v$ is the $r$-th smallest of their magnitudes and $q_i^v$ is the $r$-th quantile of $\chi_d$. We set the ``target'' magnitude for $u_i^v$ to
\beq{
    \forall i,v:\ \r_i^{v*} = \r_i^v + \eta_\r (q_i^v - \r_i^v),
    \label{eq:moving_magnitude}
}
where $\eta_\r$ is a step size. The rationale behind this is that minimizing the alignment term in \cref{eq:lejepa} reduces the magnitudes of the embeddings (see \cref{sec:dependence}).
Therefore, in practice, we use $\eta_\r > 1$ to keep the target magnitude past the target quantile $q_i^v$.

\paragraph{Moving the embedding directions.}
Inner products cannot be moved to their target quantiles individually, since the direction of the mean of each image $\hat m_i$ affects $\bb - 1$ inner products in \cref{eq:angular}.
In general, when $\bb > 2d - 1$, there might not be a set of directions $\hat m_1, \dots, \hat m_\bb$ with inner products all equal to their respective quantiles because there are $\bb(\bb-1)/2$ inner products to match and only $\bb(d-1)$ free coordinates for the directions.

Suppose $s_{ij}$ be the $r$-th smallest inner product and $q_{ij}$ be the $r$-th quantile of $\cosd$.
We want to find vectors $\D_i \in \reals^d$ to ensure that the new directions
\beq{
    \forall i:\ \hat m_i^* \propto \hat m_i + \eta_{\hat u} \D_i
    \label{eq:moving_direction}
}
have inner products $s_{ij}^* = \inner{\hat m_i^*}{\hat m_j^*}$ that minimize \cref{eq:angular}. Again we have used a step size $\eta_{\hat u}$ to provide us a tuning knob to balance the angular penalty against the alignment term.
Since $\hat m_i$ and $\hat m_i^*$ are unit vectors that lie on a sphere, the step $\D_i$ must be tangent to the sphere at $\hat m_i$, that is, $\D_i \perp \hat m_i$.
We show in \cref{app:ot_construction} how to exploit the Riemannian structure of the sphere to choose $\D_i = \a g_i$, where $g_i$ is proportional to the descent direction of the angular penalty on the sphere and $\a$ is a scalar that depends on the Hessian of \cref{eq:angular}. \Cref{prop:ot_step}b shows how this step is independent of the batch size, embedding dimension and scale of the embeddings. It also discusses how to compute such an $\a$ efficiently.

\paragraph{Simplifying the regularizer to PolarReg.}
Now that we know the target magnitude $\rho_i^{v*}$ of each view and the target direction $\hat m_i^*$ of each image, we can find a target embedding for each view
\[
    u_i^{v*} = \r_i^{v*} \rbr{\f{u_i^v + d_i}{\norm{u_i^v + d_i}}},
\]
where adding $d_i = \norm{m_i} \hat m_i^* - m_i$ to every view of image $i$ moves the direction of their mean embedding from $\hat m_i$ to $\hat m_i^*$, without changing its magnitude $\norm{m_i}$.
It also maintains the offset of each view from the mean.
Altogether, $\Om_\text{PolarReg-naive}$ is replaced by
\beq{
    \Om_{\text{PolarReg}} = \frac{1}{\bb \vv d} \sum_{i=1}^{\bb} \sum_{v=1}^{\vv} \norm{u_i^v - \text{stop-grad}\rbr{u_i^{v*}}}^2,
    \label{eq:ot_loss}
}
where $\text{stop-grad}$ means that no gradient is backpropagated through the target $u_i^{v*}$. Recall that $\Om_\text{PolarReg-naive}$ sets $\l_\perp = 0$ in \cref{eq:full_decomposition} and does not use the independence penalty.

\Cref{prop:ot_step}c shows that, to first order in the difference $u_i^v - u_i^{v*}$ between each embedding and its target, the gradient of $\Om_{\text{PolarReg}}$ is the same as that of $\Om_{\text{PolarReg-naive}}$.
It also discusses how, if PolarReg-naive were implemented as written, a gradient step on $\ell_{\hat u}$ in \cref{eq:angular} also changes the magnitudes of the embeddings. This makes choosing the coefficients $\l_\r$ and $\l_{\hat u}$ more difficult.
By exploiting the Riemannian structure in \cref{eq:moving_direction}, we decouple the two step sizes; $\eta_\r$ only changes the target magnitudes, and $\eta_{\hat u}$ the target directions.
\Cref{prop:ot_step}d shows that the gradient of $\Om_{\text{PolarReg}}$ is zero if and only if the gradients of both $\ell_\r$ and $\ell_{\hat u}$ are zero.
\Cref{prop:ot_step}e shows that its gradient is also bounded by the magnitudes of the mean embeddings and absolute differences between the magnitudes and their target quantiles.
The gradient of $\ell_{\hat u}$ in PolarReg-naive instead scales with $1/\norm{m_i}$.
Hence, $\Om_{\text{PolarReg}}$ needs no gradient clipping (\cref{app:trust_region}), and, unlike coefficients $\l_\r$ and $\l_{\hat u}$ in \cref{eq:full_decomposition}, the step sizes $\eta_\r$ and $\eta_{\hat u}$ do not need to change with batch size, embedding dimension or magnitudes of the embeddings.
\textbf{We set $\eta_{\hat u} = 2$ and $\eta_\r = 3$ for all experiments in this paper. Therefore PolarReg is robust to hyper-parameters.}
\Cref{app:eta} shows that, with these fixed step sizes, PolarReg outperforms PolarReg-naive in every setting, and that its accuracy changes little when either step size changes.

\section{Experimental Results}
\label{sec:results}

In this section, we discuss results obtained using PolarReg on a number of benchmark datasets.
Additionally, we show that PolarReg can be easily used for multi-modal learning and that the embedding magnitude can be used as a confidence score and to reliably detect out-of-domain images.

\subsection{PolarReg learns better features than sliced objectives in 4.7$\times$ fewer epochs}
\label{sec:res:accuracy}

\Cref{tab:independence} showed that on ImageNet-100, PolarReg-naive obtains a higher accuracy than SIGReg (81.00\% against 78.92\%) and that it reaches the final accuracy of SIGReg in 1.5$\times$ fewer epochs.
In the same setting, PolarReg (not in table) reaches a final accuracy of 82.42\% and matches the final accuracy of SIGReg with 1.8$\times$ fewer epochs.

On larger datasets like ImageNet-1k, PolarReg reaches the linear probing accuracy of sliced objectives with about 4.7$\times$ fewer epochs (see \cref{tab:in1k}).
We train ViT-B and L architectures with the same training hyper-parameters as VISReg; \cref{app:training} provides training and evaluation details.
A ViT-B encoder trained with PolarReg for 100 epochs performs about the same as VISReg trained for 400 epochs. It exceeds LeJEPA trained for 100 epochs by 3.4\%.
A ViT-L encoder trained with PolarReg for 100 epochs exceeds VISReg trained for 400 epochs, and is 1.6\% above both VISReg and LeJEPA trained for 100 epochs.
When trained for 400 epochs it is 1.8\% above VISReg trained for 400 epochs.
PolarReg shows consistent scaling behavior across model sizes (ViT-B to L) and training length (100 to 400 epochs).

\begin{figure}[t]
    \centering
    \includegraphics[width=0.85\linewidth]{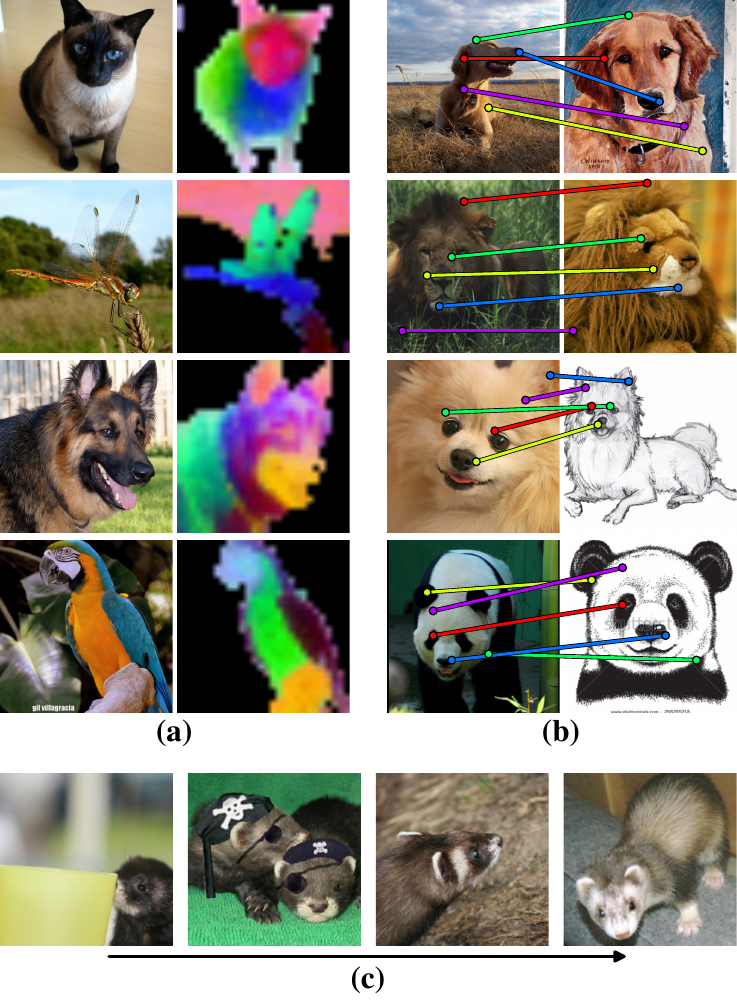}
    \vspace{0.25ex}
    \begin{adjustbox}{width=0.95\linewidth}
    \begin{tabular}{@{}llrrrrr}
        \cmidrule[\heavyrulewidth]{2-7}
        {\fontsize{11.2}{13}\selectfont\textbf{(d)}} & Top embedding magnitude (\%) & 100 & 70 & 50 & 30 & 10 \\
        \cmidrule[\lightrulewidth]{2-7}
        & Probing accuracy (\%) & 76.7 & 82.2 & 85.3 & 88.5 & 91.8 \\
        \cmidrule[\heavyrulewidth]{2-7}
    \end{tabular}
    \end{adjustbox}
    \caption{\textbf{Qualitative analysis of the embeddings of a ViT-B encoder pretrained on ImageNet-1k for 400 epochs using PolarReg.}
    \textbf{(a)} The first three principal components of the patch features, colored as RGB, separate object parts.
    \textbf{(b)} Five of the strongest matches between patch features of a photo (left) and a variant from ImageNet-R~\cite{hendrycks2021many} (right).
    The encoder was pretrained only on photos like those on the left, from ImageNet-1k.
    \textbf{(c)} Images of black-footed ferrets, sorted by increasing embedding magnitude. Their view disagreement, averaged over draws of 10 augmented views, falls from left to right; the ferrets also look less occluded or ambiguous.
    \textbf{(d)} Linear probing accuracy on ImageNet-1k is higher on images with larger embedding magnitudes.
    }
    \label{fig:results}
\end{figure}

Features of PolarReg encoders on ImageNet-1k transfer quite well to other datasets in \cref{tab:in1k}.
In particular, the ViT-B encoder trained with PolarReg for 400 epochs transfers much better than the VISReg or LeJEPA encoders. It is on par with modern SSL methods on ImageNet to CIFAR transfer evaluation.
\Cref{fig:results} qualitatively analyzes PolarReg features. It shows that they are semantically consistent, and illustrates that the embedding magnitude is smaller for semantically ambiguous images.

\paragraph{PolarReg generalizes across domains, tasks and dataset sizes.}
Features of PolarReg encoders on ImageNet-1k transfer to six datasets from other domains, such as galaxies, chest X-rays and abdominal CT, and to tasks like semantic segmentation of ADE20K (\cref{tab:app:ood}).
For these tasks, PolarReg performs similar to most other SSL methods, including the sliced objectives.
Compared to other methods, PolarReg is also better at pretraining small models on small datasets.
A ViT-S/8 encoder pretrained using PolarReg on Galaxy10, with about 16,000 training images, outperforms all other methods in \cref{tab:app:galaxy} at the end of training, and matches other methods with 2$\times$ fewer epochs.

\begin{table*}[t]
    \centering
    \caption{\textbf{PolarReg reaches the classification accuracy of sliced objectives with a quarter of the epochs.}
    Linear probing accuracy (\%) on ImageNet-1k and smaller datasets, using frozen features of encoders pretrained on ImageNet-1k.
    The classifier is trained for 100 epochs on ImageNet-1k and 10 epochs on other datasets (details in \cref{app:training}).
    $^\dagger$ Denotes that the classifier is trained for 100 epochs for all datasets.
    }
    \label{tab:in1k}
    \small
    \begin{minipage}{0.7\linewidth}
    \begin{adjustbox}{max width=\linewidth}
    \renewcommand{\arraystretch}{1.25}
    \begin{tabular}{|llr|rrrrrrrr|}
        \toprule
        Method & Backbone & Epochs & ImageNet-1k & Aircraft & Cars & CIFAR-10 & CIFAR-100 & Flowers & Food & Pets \\
        \midrule
        MoCo v3~\cite{chen2021mocov3} & ViT-B/16 & 300 & 76.7 & 57.9 & 67.5 & 96.9 & 85.2 & 91.5 & 81.8 & 89.8 \\
        DINO~\cite{caron2021dino} & ViT-B/16 & 400 & 78.2 & 63.6 & 73.9 & 96.5 & 85.0 & 94.6 & 83.1 & 93.6 \\
        iBOT~\cite{zhou2022ibot} & ViT-B/16 & 400 & 79.8 & 63.5 & 73.8 & 97.1 & 85.9 & 93.7 & 84.2 & 93.6 \\
        I-JEPA~\cite{assran2023ijepa} & ViT-H/14 & 300 & 79.3 & 55.4 & 59.2 & 97.2 & 85.5 & 86.8 & 83.3 & 92.8 \\
        \midrule
        MAE~\cite{he2022mae} & ViT-L/16 & 1600 & 75.1 & 61.9 & 61.5 & 93.3 & 78.0 & 85.4 & 78.6 & 91.3 \\
        VISReg & ViT-B/16 & 400 & 75.7 & 57.1 & 64.8 & 94.6 & 78.8 & 90.4 & 82.9 & 88.3 \\
        VISReg & ViT-L/14 & 400 & 77.0 & 56.6 & 66.2 & 94.1 & 71.9 & 90.2 & 83.3 & 89.2 \\
        LeJEPA (ours) & ViT-B/16 & 100 & 72.1 & 54.9 & 55.3 & 94.7 & 80.6 & 87.8 & 77.2 & 82.9 \\
        LeJEPA$^\dagger$ & ViT-L/14 & 100 & 75.6 & 57.0 & 57.3 & 96.5 & 83.7 & 91.2 & 82.1 & 89.7 \\
        VISReg$^\dagger$ & ViT-L/14 & 100 & 75.6 & 57.8 & 66.8 & 95.9 & 84.2 & 92.3 & 83.9 & 88.7 \\
        \midrule
        PolarReg-naive & ViT-B/16 & 100 & 74.5 & 55.7 & 62.9 & 95.1 & 82.4 & 87.0 & 79.3 & 85.6 \\
        PolarReg & ViT-B/16 & 100 & 75.5 & 57.3 & 64.5 & 94.5 & 83.7 & 87.8 & 81.1 & 89.0 \\
        PolarReg & ViT-B/16 & 400 & 76.7 & 57.9 & 68.1 & 97.0 & 85.4 & 90.6 & 82.7 & 92.7 \\
        PolarReg$^\dagger$ & ViT-L/14 & 100 & 77.2 & 57.0 & 66.3 & 96.6 & 84.8 & 86.2 & 83.8 & 92.5 \\
        PolarReg & ViT-L/14 & 400 & 78.8 & 57.3 & 67.9 & 97.2 & 85.7 & 92.0 & 83.1 & 91.5 \\
        \bottomrule
    \end{tabular}
    \end{adjustbox}
    \end{minipage}%
    \hfill
    \begin{minipage}{0.27\linewidth}
        \centering
        \captionsetup{justification=raggedright,singlelinecheck=false}
        \includegraphics[width=\linewidth]{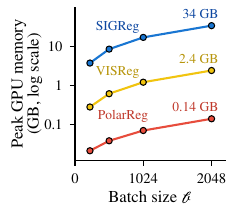}
        \captionof{figure}{\textbf{Peak GPU memory usage of the regularizers} in a forward and backward pass.}
        \label{fig:memory}
    \end{minipage}
\end{table*}

\paragraph{PolarReg is more memory and compute efficient than sliced objectives.}
On a single RTX 4090, computing PolarReg takes $\sim$0.52 ms for batch sizes ranging from 224 to 1024. It is 4--18$\times$ faster than VISReg and 35--164$\times$ faster than SIGReg, both computed with 4096 slices. See \cref{app:cost} for details.
The computation time of PolarReg is bound by GPU kernel launch overhead, whereas sliced objectives are memory-bound with latency that scales with $\bb$.
As shown in \cref{fig:memory}, PolarReg forward and backward passes need much less GPU memory, about 0.14 GB at $\bb = 2048$ against 2.4 GB for VISReg and 34 GB for SIGReg.

\subsection{PolarReg generalizes to other modalities, and classifiers trained on one modality can transfer to others}
\label{sec:res:multimodal}

PolarReg can be readily used for other modalities, such as depth images, with the same objective and hyper-parameters.
We train one ViT-S/8 encoder per modality on RGB and depth images of TartanAir v2~\cite{patel2025tartanground}, using only daytime data, and test on both daytime and nighttime data.
A linear classifier on frozen features is used to predict the scene (out of 18 scenes) given the RGB or depth image.
See \cref{app:multimodal} for details.
\Cref{fig:app:multimodal:table} shows that the PolarReg depth encoder achieves a classification accuracy of 79.3\% on daytime data as opposed to 95.9\% for the RGB data.
At night, the RGB classifier falls to 30.3\%, while the depth classifier is only mildly affected, with an accuracy of 71.6\%.

In \cref{fig:app:multimodal:table}, the two encoders are trained separately, so their features are not aligned. As expected, classifiers trained on RGB encoders perform poorly on depth features.
An orthogonal map, fitted on just 8192 unlabeled pairs of daytime RGB and depth samples, can be used to align the features.
After doing so, the RGB classifier is almost as accurate on depth encoder features as the depth classifier (75.1\% against 79.3\% at day, 68.5\% against 71.6\% at night), without using any depth labels.
This shows that separately trained encoders can be aligned with just a few unlabeled pairs. Using PolarReg, one can transfer a classifier trained on one modality to another without ground truth labels.

\subsection{The embedding magnitude detects out-of-domain images and acts as a confidence score for classification}
\label{sec:res:radius}

In \cref{fig:results}c, the ferrets look less ambiguous and have lower view disagreement with rising embedding magnitude.
This is demonstrated in \cref{fig:shape} and explained further in \cref{prop:rearrangement}, suggesting that the embedding magnitude can be a label-free measure of the semantic ambiguity of an image.

Since images with higher view disagreement get smaller magnitudes, most images unlike the training data should fall in the lower tail of $\chi_d$.
We test this on a ViT-B trained with PolarReg on ImageNet-1k by thresholding the magnitude. We report the AUROC of the fraction of test images it rejects against the fraction of ImageNet-1k validation images it rejects. See \cref{app:radius} for details.
\Cref{tab:app:radius} shows that this detects heavy corruptions (AUROC 0.99 for $8 \times 8$ block averaging) and out-of-domain images such as galaxies, chest X-rays and retina scans (0.80--0.95).
It misses mild blur and contrast changes, to which augmentations make the encoder invariant. This test needs no calibration data. The threshold that rejects a given fraction of in-domain images should be a quantile of $\chi_d$, known before training.

Images whose augmented views disagree more are often ambiguous, for instance when an object is small, occluded or placed among other objects.
We expect images with larger embedding magnitudes to be classified more accurately.
We sort the ImageNet-1k validation images by their embedding magnitudes and measure accuracy on different fractions of images with the largest magnitudes. As \cref{fig:results}d shows, keeping fewer images, from all of them to the 10\% with the largest magnitude, accuracy of the probe rises monotonically from 76.7\% to 91.8\%. Hence, the probe is more accurate on images with larger embedding magnitudes.

\clearpage

\section*{LLM Usage Statement}
We used Claude Code, a coding agent built by Anthropic.
The authors wrote the text. The tool proposed edits for grammar, and checked the notation and terms for consistency. The authors reviewed and edited every sentence.
The tool implemented and modified scripts for the figures, and parts of the training, evaluation and timing code, following the authors' specifications. The authors checked every figure and reported evaluation.
The tool checked the proofs in the supplementary material, both analytically and numerically. The authors verified every proof.
The research questions, methods and conclusions are the authors' own. No other AI tools were used.
The authors take full responsibility for all content in this paper.
{
    \small
    \bibliographystyle{ieeenat_fullname}
    \bibliography{bib/spolar}

\begin{thebibliography}{48}
\providecommand{\natexlab}[1]{#1}
\providecommand{\url}[1]{\texttt{#1}}
\expandafter\ifx\csname urlstyle\endcsname\relax
  \providecommand{\doi}[1]{doi: #1}\else
  \providecommand{\doi}{doi: \begingroup \urlstyle{rm}\Url}\fi

\bibitem[Agrawal et~al.(2022)Agrawal, Mondal, Ghosh, and
  Richards]{agrawal2022alphareq}
Kumar~K Agrawal, Arnab~Kumar Mondal, Arna Ghosh, and Blake Richards.
\newblock {$\alpha$-ReQ}: Assessing representation quality in self-supervised
  learning by measuring eigenspectrum decay.
\newblock In \emph{Advances in Neural Information Processing Systems}, pages
  17626--17638, 2022.

\bibitem[Ansel et~al.(2024)Ansel, Yang, He, Gimelshein, Jain, Voznesensky, Bao,
  Bell, Berard, Burovski, Chauhan, Chourdia, Constable, Desmaison, DeVito,
  Ellison, Feng, Gong, Gschwind, Hirsh, Huang, Kalambarkar, Kirsch, Lazos,
  Lezcano, Liang, Liang, Lu, Luk, Maher, Pan, Puhrsch, Reso, Saroufim,
  Siraichi, Suk, Zhang, Suo, Tillet, Zhao, Wang, Zhou, Zou, Wang, Mathews, Wen,
  Chanan, Wu, and Chintala]{ansel2024pytorch2}
Jason Ansel, Edward Yang, Horace He, Natalia Gimelshein, Animesh Jain, Michael
  Voznesensky, Bin Bao, Peter Bell, David Berard, Evgeni Burovski, Geeta
  Chauhan, Anjali Chourdia, Will Constable, Alban Desmaison, Zachary DeVito,
  Elias Ellison, Will Feng, Jiong Gong, Michael Gschwind, Brian Hirsh, Sherlock
  Huang, Kshiteej Kalambarkar, Laurent Kirsch, Michael Lazos, Mario Lezcano,
  Yanbo Liang, Jason Liang, Yinghai Lu, C.~K. Luk, Bert Maher, Yunjie Pan,
  Christian Puhrsch, Matthias Reso, Mark Saroufim, Marcos~Yukio Siraichi, Helen
  Suk, Shunting Zhang, Michael Suo, Phil Tillet, Xu Zhao, Eikan Wang, Keren
  Zhou, Richard Zou, Xiaodong Wang, Ajit Mathews, William Wen, Gregory Chanan,
  Peng Wu, and Soumith Chintala.
\newblock {PyTorch} 2: Faster machine learning through dynamic {Python}
  bytecode transformation and graph compilation.
\newblock In \emph{Proceedings of the 29th ACM International Conference on
  Architectural Support for Programming Languages and Operating Systems, Volume
  2}, pages 929--947, New York, NY, USA, 2024. Association for Computing
  Machinery.

\bibitem[Assran et~al.(2023)Assran, Duval, Misra, Bojanowski, Vincent, Rabbat,
  LeCun, and Ballas]{assran2023ijepa}
Mahmoud Assran, Quentin Duval, Ishan Misra, Piotr Bojanowski, Pascal Vincent,
  Michael Rabbat, Yann LeCun, and Nicolas Ballas.
\newblock Self-supervised learning from images with a joint-embedding
  predictive architecture.
\newblock In \emph{IEEE/CVF Conference on Computer Vision and Pattern
  Recognition (CVPR)}, 2023.

\bibitem[Bai et~al.(2026)Bai, Gong, Liu, Weng, Wen, and Yap]{bai2026mmlejepa}
Haowen Bai, Rui Gong, Meichen Liu, Ming-Xuan Weng, Bihan Wen, and Kim-Hui Yap.
\newblock {MM-LeJEPA}: Correlated {Gaussian} regularization for multimodal
  joint-embedding predictive architectures.
\newblock \emph{arXiv preprint arXiv:2610.24681}, 2026.

\bibitem[Balestriero and LeCun(2025)]{lejepa2025}
Randall Balestriero and Yann LeCun.
\newblock {LeJEPA}: Provable and scalable self-supervised learning without the
  heuristics.
\newblock \emph{arXiv preprint arXiv:2511.08544}, 2025.

\bibitem[Bardes et~al.(2022)Bardes, Ponce, and LeCun]{bardes2022vicreg}
Adrien Bardes, Jean Ponce, and Yann LeCun.
\newblock {VICReg}: Variance-invariance-covariance regularization for
  self-supervised learning.
\newblock In \emph{International Conference on Learning Representations
  (ICLR)}, 2022.

\bibitem[Bobkov and Ledoux(2019)]{bobkov2019one}
Sergey Bobkov and Michel Ledoux.
\newblock One-dimensional empirical measures, order statistics, and
  {K}antorovich transport distances.
\newblock \emph{Memoirs of the American Mathematical Society}, 261\penalty0
  (1259), 2019.

\bibitem[Bonet et~al.(2023)Bonet, Berg, Courty, Septier, Drumetz, and
  Pham]{bonet2023ssw}
Cl{\'e}ment Bonet, Paul Berg, Nicolas Courty, Fran{\c{c}}ois Septier, Lucas
  Drumetz, and Minh-Tan Pham.
\newblock Spherical {Sliced-Wasserstein}.
\newblock In \emph{International Conference on Learning Representations
  (ICLR)}, 2023.

\bibitem[Bonneel et~al.(2015)Bonneel, Rabin, Peyr{\'e}, and
  Pfister]{bonneel2015sliced}
Nicolas Bonneel, Julien Rabin, Gabriel Peyr{\'e}, and Hanspeter Pfister.
\newblock Sliced and {R}adon {W}asserstein barycenters of measures.
\newblock \emph{Journal of Mathematical Imaging and Vision}, 51\penalty0
  (1):\penalty0 22--45, 2015.

\bibitem[Boumal(2023)]{boumal2023introduction}
Nicolas Boumal.
\newblock \emph{An Introduction to Optimization on Smooth Manifolds}.
\newblock Cambridge University Press, 2023.

\bibitem[Caron et~al.(2021)Caron, Touvron, Misra, J{\'e}gou, Mairal,
  Bojanowski, and Joulin]{caron2021dino}
Mathilde Caron, Hugo Touvron, Ishan Misra, Herv{\'e} J{\'e}gou, Julien Mairal,
  Piotr Bojanowski, and Armand Joulin.
\newblock Emerging properties in self-supervised vision transformers.
\newblock In \emph{IEEE/CVF International Conference on Computer Vision
  (ICCV)}, 2021.

\bibitem[Chen et~al.(2021)Chen, Xie, and He]{chen2021mocov3}
Xinlei Chen, Saining Xie, and Kaiming He.
\newblock An empirical study of training self-supervised vision transformers.
\newblock In \emph{2021 IEEE/CVF International Conference on Computer Vision
  (ICCV)}, pages 9620--9629, 2021.

\bibitem[Epps and Pulley(1983)]{epps1983test}
T.~W. Epps and L.~B. Pulley.
\newblock A test for normality based on the empirical characteristic function.
\newblock \emph{Biometrika}, 70\penalty0 (3):\penalty0 723--726, 1983.

\bibitem[Ermolov et~al.(2021)Ermolov, Siarohin, Sangineto, and
  Sebe]{ermolov2021wmse}
Aleksandr Ermolov, Aliaksandr Siarohin, Enver Sangineto, and Nicu Sebe.
\newblock Whitening for self-supervised representation learning.
\newblock In \emph{Proceedings of the 38th International Conference on Machine
  Learning}, pages 3015--3024. PMLR, 2021.

\bibitem[Fang et~al.(2024)Fang, Li, Sun, and Wang]{fang2024uniformity}
Xianghong Fang, Jian Li, Qiang Sun, and Benyou Wang.
\newblock Rethinking the uniformity metric in self-supervised learning.
\newblock In \emph{International Conference on Learning Representations
  (ICLR)}, 2024.

\bibitem[Feuerverger and Mureika(1977)]{feuerverger1977ecf}
Andrey Feuerverger and Roman~A. Mureika.
\newblock The empirical characteristic function and its applications.
\newblock \emph{The Annals of Statistics}, 5\penalty0 (1):\penalty0 88--97,
  1977.

\bibitem[Garrido et~al.(2023)Garrido, Balestriero, Najman, and
  LeCun]{garrido2023rankme}
Quentin Garrido, Randall Balestriero, Laurent Najman, and Yann LeCun.
\newblock {R}ank{M}e: Assessing the downstream performance of pretrained
  self-supervised representations by their rank.
\newblock In \emph{Proceedings of the 40th International Conference on Machine
  Learning}, pages 10929--10974. PMLR, 2023.

\bibitem[Gretton et~al.(2005)Gretton, Bousquet, Smola, and
  Sch{\"o}lkopf]{gretton2005hsic}
Arthur Gretton, Olivier Bousquet, Alex Smola, and Bernhard Sch{\"o}lkopf.
\newblock Measuring statistical dependence with {Hilbert-Schmidt} norms.
\newblock In \emph{International Conference on Algorithmic Learning Theory
  (ALT)}, 2005.

\bibitem[Gretton et~al.(2012)Gretton, Borgwardt, Rasch, Sch{{\"o}}lkopf, and
  Smola]{JMLR:v13:gretton12a}
Arthur Gretton, Karsten~M. Borgwardt, Malte~J. Rasch, Bernhard Sch{{\"o}}lkopf,
  and Alexander Smola.
\newblock A kernel two-sample test.
\newblock \emph{Journal of Machine Learning Research}, 13\penalty0
  (25):\penalty0 723--773, 2012.

\bibitem[Hardy et~al.(1934)Hardy, Littlewood, and
  P{\'o}lya]{hardy1934inequalities}
G.~H. Hardy, J.~E. Littlewood, and G. P{\'o}lya.
\newblock \emph{Inequalities}.
\newblock Cambridge University Press, 1934.

\bibitem[He et~al.(2022)He, Chen, Xie, Li, Doll{\'a}r, and Girshick]{he2022mae}
Kaiming He, Xinlei Chen, Saining Xie, Yanghao Li, Piotr Doll{\'a}r, and Ross
  Girshick.
\newblock Masked autoencoders are scalable vision learners.
\newblock In \emph{IEEE/CVF Conference on Computer Vision and Pattern
  Recognition (CVPR)}, 2022.

\bibitem[Hendrycks et~al.(2021)Hendrycks, Basart, Mu, Kadavath, Wang, Dorundo,
  Desai, Zhu, Parajuli, Guo, Song, Steinhardt, and Gilmer]{hendrycks2021many}
Dan Hendrycks, Steven Basart, Norman Mu, Saurav Kadavath, Frank Wang, Evan
  Dorundo, Rahul Desai, Tyler Zhu, Samyak Parajuli, Mike Guo, Dawn Song, Jacob
  Steinhardt, and Justin Gilmer.
\newblock The many faces of robustness: A critical analysis of
  out-of-distribution generalization.
\newblock In \emph{IEEE/CVF International Conference on Computer Vision
  (ICCV)}, 2021.

\bibitem[Hua et~al.(2021)Hua, Wang, Xue, Ren, Wang, and
  Zhao]{hua2021decorrelation}
Tianyu Hua, Wenxiao Wang, Zihui Xue, Sucheng Ren, Yue Wang, and Hang Zhao.
\newblock On feature decorrelation in self-supervised learning.
\newblock In \emph{Proceedings of the IEEE/CVF International Conference on
  Computer Vision (ICCV)}, pages 9598--9608, 2021.

\bibitem[Jiang and Pham(2025)]{jiang2025hyperspheres}
Tiefeng Jiang and Tuan Pham.
\newblock Detecting non-uniform patterns on high-dimensional hyperspheres.
\newblock \emph{arXiv preprint arXiv:2506.00444}, 2025.

\bibitem[Jing et~al.(2022)Jing, Vincent, LeCun, and Tian]{jing2022dimensional}
Li Jing, Pascal Vincent, Yann LeCun, and Yuandong Tian.
\newblock Understanding dimensional collapse in contrastive self-supervised
  learning.
\newblock In \emph{International Conference on Learning Representations
  (ICLR)}, 2022.

\bibitem[Le(2026)]{urjepa2026}
Triet~M. Le.
\newblock {UR-JEPA}: Uniform rectifiability as a regularizer for
  joint-embedding predictive architectures.
\newblock \emph{arXiv preprint arXiv:2606.01443}, 2026.

\bibitem[LeCun(2022)]{lecun2022path}
Yann LeCun.
\newblock A path towards autonomous machine intelligence.
\newblock OpenReview, 2022.

\bibitem[Li et~al.(2022)Li, Efros, and Pathak]{li2022collapse}
Alexander~C. Li, Alexei~A. Efros, and Deepak Pathak.
\newblock Understanding collapse in non-contrastive {Siamese} representation
  learning.
\newblock In \emph{Computer Vision – ECCV 2022}, pages 490--505. Springer
  Nature Switzerland, 2022.

\bibitem[Mordacq et~al.(2025)Mordacq, Loiseaux, Kalogeiton, and
  Oudot]{mordacq2025tregs}
Julie Mordacq, David Loiseaux, Vicky Kalogeiton, and Steve Oudot.
\newblock {T-REGS}: Minimum spanning tree regularization for self-supervised
  learning.
\newblock In \emph{Advances in Neural Information Processing Systems}, pages
  12184--12214, 2025.

\bibitem[Muirhead(1982)]{muirhead1982aspects}
Robb~J. Muirhead.
\newblock \emph{Aspects of Multivariate Statistical Theory}.
\newblock John Wiley \& Sons, 1982.

\bibitem[Nicollier et~al.(2026)Nicollier, Dunitz, Pic, Mus{é},
  Meinhardt-Llopis, and Facciolo]{nicollier2026spherejepa}
L{é}o Nicollier, Max Dunitz, Marc Pic, Pablo Mus{é}, Enric Meinhardt-Llopis,
  and Gabriele Facciolo.
\newblock {SPHERE-JEPA}: Spherical prediction with homogeneous embeddings.
\newblock \emph{arXiv preprint arXiv:2605.26900}, 2026.

\bibitem[Oquab et~al.(2024)Oquab, Darcet, Moutakanni, Vo, Szafraniec, Khalidov,
  Fernandez, Haziza, Massa, El-Nouby, Assran, Ballas, Galuba, Howes, Huang, Li,
  Misra, Rabbat, Sharma, Synnaeve, Xu, J{\'e}gou, Mairal, Labatut, Joulin, and
  Bojanowski]{oquab2024dinov2}
Maxime Oquab, Timoth{\'e}e Darcet, Th{\'e}o Moutakanni, Huy~V. Vo, Marc
  Szafraniec, Vasil Khalidov, Pierre Fernandez, Daniel Haziza, Francisco Massa,
  Alaaeldin El-Nouby, Mahmoud Assran, Nicolas Ballas, Wojciech Galuba, Russell
  Howes, Po-Yao Huang, Shang-Wen Li, Ishan Misra, Michael Rabbat, Vasu Sharma,
  Gabriel Synnaeve, Hu Xu, Herv{\'e} J{\'e}gou, Julien Mairal, Patrick Labatut,
  Armand Joulin, and Piotr Bojanowski.
\newblock {DINO}v2: Learning robust visual features without supervision.
\newblock \emph{Transactions on Machine Learning Research}, 2024.

\bibitem[Patel et~al.(2025)Patel, Yang, Qiu, Cadena, Scherer, Hutter, and
  Wang]{patel2025tartanground}
Manthan Patel, Fan Yang, Yuheng Qiu, Cesar Cadena, Sebastian Scherer, Marco
  Hutter, and Wenshan Wang.
\newblock {TartanGround}: A large-scale dataset for ground robot perception and
  navigation.
\newblock In \emph{2025 IEEE/RSJ International Conference on Intelligent Robots
  and Systems (IROS)}, pages 20524--20531, 2025.

\bibitem[Ramesh et~al.(2025)Ramesh, Bisulco, DiTullio, Wei, Balasubramanian,
  Daniilidis, and Chaudhari]{ramesh2024many}
Rahul Ramesh, Anthony Bisulco, Ronald~W. DiTullio, Linran Wei, Vijay
  Balasubramanian, Kostas Daniilidis, and Pratik Chaudhari.
\newblock Many perception tasks are highly redundant functions of their input
  data.
\newblock In \emph{Computational and {{Systems Neuroscience}} ({{COSYNE}})},
  2025.

\bibitem[Seetharaman et~al.(2020)Seetharaman, Wichern, Pardo, and
  Le~Roux]{seetharaman2020autoclip}
Prem Seetharaman, Gordon Wichern, Bryan Pardo, and Jonathan Le~Roux.
\newblock {AutoClip}: Adaptive gradient clipping for source separation
  networks.
\newblock In \emph{IEEE International Workshop on Machine Learning for Signal
  Processing (MLSP)}, 2020.

\bibitem[Shazeer and Stern(2018)]{shazeer2018adafactor}
Noam Shazeer and Mitchell Stern.
\newblock Adafactor: Adaptive learning rates with sublinear memory cost.
\newblock In \emph{Proceedings of the 35th International Conference on Machine
  Learning}, pages 4596--4604, 2018.

\bibitem[Song et~al.(2012)Song, Smola, Gretton, Bedo, and
  Borgwardt]{song2012feature}
Le Song, Alex Smola, Arthur Gretton, Justin Bedo, and Karsten Borgwardt.
\newblock Feature selection via dependence maximization.
\newblock \emph{Journal of Machine Learning Research}, 13\penalty0
  (47):\penalty0 1393--1434, 2012.

\bibitem[Tandon et~al.(2026)Tandon, Borgi, Snopov,
  Garc{\'{\symbol{16}}}a-Redondo, D'Ascenzo, di~Montesano, Silvestri, Liò,
  Dominici, Balestriero, and Battiloro]{tandon2026localejepa}
Kartik Tandon, Alessio Borgi, Pavel Snopov, Inés
  Garc{\'{\symbol{16}}}a-Redondo, Davide D'Ascenzo, Sebastiano~Cultrera di
  Montesano, Fabrizio Silvestri, Pietro Liò, Francesca Dominici, Randall
  Balestriero, and Claudio Battiloro.
\newblock {LocaLeJEPA}: Maximum entropy representations under geometric priors.
\newblock \emph{arXiv preprint arXiv:2610.28458}, 2026.

\bibitem[Thilak et~al.(2024)Thilak, Huang, Saremi, Dinh, Goh, Nakkiran,
  Susskind, and Littwin]{thilak2024lidar}
Vimal Thilak, Chen Huang, Omid Saremi, Laurent Dinh, Hanlin Goh, Preetum
  Nakkiran, Joshua Susskind, and Etai Littwin.
\newblock {LiDAR}: Sensing linear probing performance in joint embedding {SSL}
  architectures.
\newblock In \emph{International Conference on Learning Representations}, pages
  56726--56765, 2024.

\bibitem[Tian et~al.(2020)Tian, Krishnan, and Isola]{tian2020cmc}
Yonglong Tian, Dilip Krishnan, and Phillip Isola.
\newblock Contrastive multiview coding.
\newblock In \emph{European Conference on Computer Vision (ECCV)}, 2020.

\bibitem[Villani(2003)]{villani2003topics}
C{\'e}dric Villani.
\newblock \emph{Topics in Optimal Transportation}.
\newblock American Mathematical Society, 2003.

\bibitem[Wang and Isola(2020)]{wang2020alignment}
Tongzhou Wang and Phillip Isola.
\newblock Understanding contrastive representation learning through alignment
  and uniformity on the hypersphere.
\newblock In \emph{Proceedings of the 37th International Conference on Machine
  Learning}, pages 9929--9939. PMLR, 2020.

\bibitem[Welch(1974)]{welch1974lower}
L.~R. Welch.
\newblock Lower bounds on the maximum cross correlation of signals.
\newblock \emph{IEEE Transactions on Information Theory}, 20\penalty0
  (3):\penalty0 397--399, 1974.

\bibitem[Weng et~al.(2024)Weng, Ni, Song, Luo, Anwer, Khan, Khan, and
  Huang]{weng2024intl}
Xi Weng, Yunhao Ni, Tengwei Song, Jie Luo, Rao~Muhammad Anwer, Salman Khan,
  Fahad~Shahbaz Khan, and Lei Huang.
\newblock Modulate your spectrum in self-supervised learning.
\newblock In \emph{International Conference on Learning Representations}, pages
  670--697, 2024.

\bibitem[Wu et~al.(2026)Wu, Balestriero, and Levine]{wu2026visreg}
Haiyu Wu, Randall Balestriero, and Morgan Levine.
\newblock {VISReg}: Variance-invariance-sketching regularization for {JEPA}
  training.
\newblock \emph{arXiv preprint arXiv:2606.02572}, 2026.

\bibitem[Zbontar et~al.(2021)Zbontar, Jing, Misra, LeCun, and
  Deny]{zbontar2021barlow}
Jure Zbontar, Li Jing, Ishan Misra, Yann LeCun, and Stephane Deny.
\newblock {Barlow Twins}: Self-supervised learning via redundancy reduction.
\newblock In \emph{Proceedings of the 38th International Conference on Machine
  Learning}, pages 12310--12320. PMLR, 2021.

\bibitem[Zhou et~al.(2022)Zhou, Wei, Wang, Shen, Xie, Yuille, and
  Kong]{zhou2022ibot}
Jinghao Zhou, Chen Wei, Huiyu Wang, Wei Shen, Cihang Xie, Alan Yuille, and Tao
  Kong.
\newblock {iBOT}: Image {BERT} pre-training with online tokenizer.
\newblock In \emph{International Conference on Learning Representations
  (ICLR)}, 2022.

\bibitem[Zimmermann et~al.(2025)Zimmermann, Wiltzer, Szeto, Alvarez-Melis, and
  Mackey]{zimmermann2025kerjepa}
Eric Zimmermann, Harley Wiltzer, Justin Szeto, David Alvarez-Melis, and Lester
  Mackey.
\newblock {KerJEPA}: Kernel discrepancies for {Euclidean} self-supervised
  learning.
\newblock \emph{arXiv preprint arXiv:2512.19605}, 2025.

\end{thebibliography}
}

\clearpage
\setcounter{page}{1}
\renewcommand{\thepage}{S\arabic{page}}

\setlength{\textwidth}{6.2in}
\setlength{\oddsidemargin}{0.15in}
\setlength{\evensidemargin}{0.15in}
\setlength{\parskip}{1ex}
\onecolumn
\begin{center}
    \Large
    \textbf{\thetitle}\\
    \vspace{0.5em}Supplementary Material\\
    \vspace{1.0em}
\end{center}

\appendix
\counterwithin{theorem}{section}
\counterwithin{proposition}{section}
\counterwithin{lemma}{section}
\counterwithin{corollary}{section}
\counterwithin{assumption}{section}
\counterwithin{definition}{section}
\counterwithin{example}{section}
\counterwithin{remark}{section}
\counterwithin{fact}{section}
\counterwithin{table}{section}
\counterwithin{figure}{section}

\section{Related work}
\label{app:related}

Several variants of LeJEPA~\cite{lejepa2025} change its regularizer or its target.
VISReg~\cite{wu2026visreg} matches the sliced projections of SIGReg with a Wasserstein distance.
KerJEPA~\cite{zimmermann2025kerjepa} replaces the sliced projections with a kernel discrepancy between the embeddings and the target.
SPHERE-JEPA~\cite{nicollier2026spherejepa} normalizes the embeddings and changes the target of SIGReg to the uniform distribution on the sphere.
UR-JEPA~\cite{urjepa2026} changes the target to distributions that concentrate near a low-dimensional subset of the embedding space.
Others change the target when domain-specific geometric priors are available~\cite{tandon2026localejepa} or to handle multiple modalities~\cite{bai2026mmlejepa}.
We enforce the marginal distributions of an isotropic Gaussian for the magnitude and the direction of the embeddings, without requiring the two to be independent.
Our angular penalty is related to losses that make the embedding directions uniform on a sphere~\cite{wang2020alignment,bonet2023ssw,fang2024uniformity,mordacq2025tregs}.
A regularizer must also prevent dimensional collapse, where the embeddings end up only in a lower-dimensional subspace~\cite{jing2022dimensional,hua2021decorrelation,li2022collapse}. Such collapse is often measured by the rank or the spectrum of the embeddings~\cite{garrido2023rankme,agrawal2022alphareq,thilak2024lidar}. Many methods prevent it by constraining the covariance of the embeddings~\cite{bardes2022vicreg,zbontar2021barlow,ermolov2021wmse,weng2024intl}. PolarReg prevents it by constraining the distributions of the magnitudes and of the pairwise inner products (\cref{rem:app:coupling}).

\section{Proofs}
\label{app:proofs}

\subsection{Expected value of SIGReg on non-isotropic Gaussian embeddings}
\label{app:sliced_power}

For each view, SIGReg tests a batch of embeddings $u_1, \dots, u_\bb$ one slice at a time.
Along a slice $a$, it compares the empirical characteristic function (ECF) of the embedding projections $a^\top u_1, \dots, a^\top u_\bb$ with $e^{-t^2/2}$, the characteristic function of $\text{N}(0, 1)$ (\cref{eq:lejepa}).
If the embeddings are samples of $\text{N}(0, \Sigma)$, their projections along a slice $a$ are samples of $\text{N}(0, a^\top \Sigma a)$.
So on a slice, the expected value of SIGReg depends on $\Sigma$ only through $\e_a = a^\top \Sigma a - 1$, which is zero for control batches.
\Cref{prop:sliced_power} computes how large $\e_a$ is on a random slice, and what this implies for the expectation.
Particularly, it formalizes why more slices do not help in differentiating between collapsed and control batches, but more batches do.

\begin{proposition}
\label{prop:sliced_power}
Let the embeddings be independent samples of $\text{N}(0, \Sigma)$ with $\operatorname{tr}(\Sigma) = d$, and let $\omega = \norm{\Sigma - I}_F$ measure how far $\Sigma$ is from isotropic.
For $\Sigma$ with $r$ equal nonzero eigenvalues, $\omega^2/d^2 = 1/r - 1/d$.
Along a uniformly random unit slice $a$, the projections of the embeddings are samples from $\text{N}(0, 1 + \e_a)$ with $\E\sbr{\e_a} = 0$ and $\E\sbr{\e_a^2} = 2\omega^2/(d(d+2))$.
SIGReg with $\kk$ slices on a batch of $\bb$ embeddings, denoted by $T_\kk$, has, up to terms of third order in $\e_a$, the expectation
\beq{
    \E\sbr{T_\kk} = c_0 + c_1 (\bb - 2)\, \frac{2\omega^2}{d(d+2)},
    \label{eq:sliced_power}
}
where $c_0$ and $c_1$ depend only on the weight $w$ in \cref{eq:lejepa}.
The expectation does not depend on $\kk$.
\end{proposition}

For $\Sigma = I$ the embeddings form a control batch and $\E[T_\kk] = c_0$, so \cref{eq:sliced_power} tells how far, on average, SIGReg on a collapsed batch sits above its value on control batches.
The test rejects the collapse when this gap exceeds the spread of $T_\kk$ on control batches.
For $\Sigma$ with $r$ equal nonzero eigenvalues, the variance of $a^\top \Sigma a$ across random slices is $\E[\e_a^2] = 2(d-r)/(r(d+2))$.
On a control batch, the empirical variance of the $\bb$ projections along a slice has variance $2/\bb$, since each squared projection has variance 2.
Since the gap is $c_1 (\bb - 2) \E\sbr{\e_a^2}$, it is large only when $2(d-r)/(r(d+2))$ is large compared with $2/\bb$.
Neither quantity depends on $\kk$; more slices only reduce the variance of $T_\kk$ across slices.
At a fixed ratio $r/d$, $2(d-r)/(r(d+2))$ falls as $1/d$, so the gap also falls with the embedding dimension.
To prove the proposition, we use two quantities: the fourth moments of a uniformly random unit vector (\cref{lem:app:slice_moments}), and the mean and variance of an ECF~\cite{feuerverger1977ecf}.

\paragraph{Gaussian embeddings and the sliced statistic.}
We assume that the embeddings are independent samples from $\text{N}(0, \Sigma)$ with $\operatorname{tr}\Sigma = d$.
The matrix $\Sigma - I$ is symmetric with $\operatorname{tr}(\Sigma - I) = 0$, and $\omega = \norm{\Sigma - I}_F$ measures how far $\Sigma$ is from isotropic.
The slices $a_1, \dots, a_\kk$ are independent, uniform on the unit sphere $S^{d-1}$, and independent of the batch.
Along a slice $a$, SIGReg computes
\beq{
    T(a) = \bb \int w(t) \abr{\varphi_a(t) - e^{-t^2/2}}^2 \dd{t},
    \qquad
    \varphi_a(t) = \frac{1}{\bb} \sum_{i=1}^{\bb} e^{\i t\, a^\top u_i},
    \label{eq:app:slice_stat}
}
where $w(t)$ is the weight function in \cref{eq:lejepa} that keeps the integral finite.
SIGReg then averages $T(a)$ over the $\kk$ slices, $T_\kk = \kk^{-1} \sum_{k=1}^{\kk} T(a_k)$, which for one view is $\Om_{\text{SIGReg}}/\l$ in \cref{eq:lejepa}.
The arguments below hold for any $w$ with $\int w(t) (1 + t^6) \dd{t} < \infty$, including the weights LeJEPA uses.

For example, if $\Sigma$ has $r$ equal nonzero eigenvalues, each is $d/r$ because $\operatorname{tr}\Sigma = d$, and $\Sigma = (d/r) P_r$, where $P_r$ is the orthogonal projection onto the $r$-dimensional subspace spanned by the corresponding eigenvectors.
Then $\Sigma - I = (d/r) P_r - I$. Since $P_r^2 = P_r$ and $\operatorname{tr} P_r = r$,
\beq{
    \omega^2 = \operatorname{tr}\rbr{(\Sigma - I)^2} = \frac{d^2}{r^2}\, r - 2\, \frac{d}{r}\, r + d = \frac{d^2}{r} - d,
    \qquad \text{so} \qquad
    \frac{\omega^2}{d^2} = \frac1r - \frac1d.
    \label{eq:app:rank_delta}
}

\begin{lemma}[Moments of a random slice]
\label{lem:app:slice_moments}
Let $a$ be uniform on $S^{d-1}$ and let $A$ be a symmetric $d \times d$ matrix.
Then
\beq{
    \E\sbr{a^\top A a} = \frac{\operatorname{tr} A}{d},
    \qquad
    \E\sbr{\rbr{a^\top A a}^2} = \frac{(\operatorname{tr} A)^2 + 2\operatorname{tr}(A^2)}{d(d+2)}.
    \label{eq:app:slice_moments}
}
\end{lemma}
\begin{proof}
Let $g \sim \text{N}(0, I_{d \times d})$.
Since $g \neq 0$ with probability one and $Hg$ has the same distribution as $g$ for every orthogonal matrix $H$, the vector $g/\norm{g}$ is uniform on $S^{d-1}$ and independent of $\norm{g}$~\cite[Theorem 1.5.6]{muirhead1982aspects}.
So we may take $a = g/\norm{g}$, and then $g^\top A g = \norm{g}^2\, a^\top A a$.
We can write $A = Q \Lambda Q^\top$ with eigenvalues $\lambda_1, \dots, \lambda_d$.
Then $z = Q^\top g \sim \text{N}(0, I_{d \times d})$ and $g^\top A g = \sum_i \lambda_i z_i^2$.
Since the $z_i$ are independent with $\E\sbr{z_i^2} = 1$ and $\E\sbr{z_i^4} = 3$,
\beqs{
    \E\sbr{g^\top A g} = \sum_i \lambda_i = \operatorname{tr} A,
    \qquad
    \E\sbr{\rbr{g^\top A g}^2} = \sum_{i \neq j} \lambda_i \lambda_j + 3 \sum_i \lambda_i^2 = (\operatorname{tr} A)^2 + 2 \operatorname{tr}(A^2).
}
The squared norm $\norm{g}^2$ has a chi-squared distribution with $d$ degrees of freedom, with mean $d$ and variance $2d$, so $\E\sbr{\norm{g}^2} = d$ and $\E\sbr{\norm{g}^4} = 2d + d^2 = d(d+2)$.
Since $\norm{g}$ and $a$ are independent, $\E\sbr{g^\top A g} = d\, \E\sbr{a^\top A a}$ and $\E\sbr{(g^\top A g)^2} = d(d+2)\, \E\sbr{(a^\top A a)^2}$, proving the lemma.
\end{proof}

\paragraph{Proof of \cref{prop:sliced_power}.}

\noindent
\textbf{On a slice, the covariance of the embeddings affects only the variance of their projections.}
Fix a slice $a$.
Since $a$ is independent of the batch, the projections $a^\top u_1, \dots, a^\top u_\bb$ are independent samples of $\text{N}(0, a^\top \Sigma a)$, and $a^\top\Sigma a = a^\top a + a^\top (\Sigma - I) a = 1 + \e_a$ with $\e_a = a^\top (\Sigma - I) a$.
The difference $\Sigma - I$ affects the slice only through $\e_a$.

\noindent
By \cref{lem:app:slice_moments} with $A = \Sigma - I$, $\operatorname{tr}(\Sigma - I) = 0$ and $\operatorname{tr}\rbr{(\Sigma - I)^2} = \omega^2$,
\beq{
    \E\sbr{\e_a} = 0,
    \qquad
    \E\sbr{\e_a^2} = \frac{2 \omega^2}{d(d+2)}.
    \label{eq:app:eps_moments}
}

\paragraph{On one slice, the expected statistic is $\bb$ times a squared bias plus sampling noise.}
We use the mean and covariance of an ECF from \cite[Section 3]{feuerverger1977ecf}, and state them here for ease of reference.
If $x_1, \dots, x_\bb$ are independent samples of a distribution with characteristic function $\psi$, and $\varphi(t) = \frac{1}{\bb} \sum_{i=1}^{\bb} e^{\i t x_i}$, then for all $t_1$ and $t_2$,
\beq{
    \E\sbr{\varphi(t_1)} = \psi(t_1),
    \qquad
    \E\sbr{\rbr{\varphi(t_1) - \psi(t_1)} \rbr{\varphi(t_2) - \psi(t_2)}} = \frac{\psi(t_1 + t_2) - \psi(t_1)\, \psi(t_2)}{\bb}.
    \label{eq:app:ecf}
}
Since $\varphi(-t)$ and $\psi(-t)$ are the complex conjugates of $\varphi(t)$ and $\psi(t)$, setting $t_1 = t$ and $t_2 = -t$ gives $\E\sbr{\abr{\varphi(t) - \psi(t)}^2} = \rbr{1 - \abr{\psi(t)}^2} / \bb$.
Given $a$, the projections have the characteristic function $\psi_a(t) = e^{-(1 + \e_a) t^2/2}$.
We can split $\varphi_a(t) - e^{-t^2/2}$ into its mean, $\psi_a(t) - e^{-t^2/2}$, and its sampling error, $\varphi_a(t) - \psi_a(t)$, which has mean zero.
Then,
\beq{
    \E\sbr{\abr{\varphi_a(t) - e^{-t^2/2}}^2 \mid a} = \rbr{e^{-(1 + \e_a) t^2/2} - e^{-t^2/2}}^2 + \frac{1 - e^{-(1 + \e_a) t^2}}{\bb},
    \label{eq:app:ecf_split}
}
since the cross term is zero under expectation.
The first term is the squared bias, how far the distribution of projections is from an isotropic Gaussian.
The second is sampling noise, that is present even when $\e_a = 0$. It shrinks as $1/\bb$.
SIGReg multiplies both by $\bb$, so the first term grows with the batch while the second stays at a fixed level,
\beq{
    \E\sbr{T(a) \mid a} = \int w(t) \sbr{\bb \rbr{e^{-(1 + \e_a) t^2/2} - e^{-t^2/2}}^2 + 1 - e^{-(1 + \e_a) t^2}} \dd{t}.
    \label{eq:app:slice_mean}
}
We expand in $\e_a$.
The bias is $e^{-t^2/2} \rbr{e^{-\e_a t^2/2} - 1} = -\frac12 \e_a t^2 e^{-t^2/2} + O(\e_a^2)$, so its square is $\frac14 \e_a^2 t^4 e^{-t^2} + O(\e_a^3)$.
The sampling term is $1 - e^{-t^2} + \e_a t^2 e^{-t^2} - \frac12 \e_a^2 t^4 e^{-t^2} + O(\e_a^3)$.
Collecting powers of $\e_a$,
\beq{
    \E\sbr{T(a) \mid a} = c_0 + c'\, \e_a + c_1 (\bb - 2)\, \e_a^2 + O\rbr{(\bb + 1) \abr{\e_a}^3},
    \label{eq:app:slice_expansion}
}
where
\beqs{
    c_0 = \int w(t) \rbr{1 - e^{-t^2}} \dd{t},
    \qquad
    c' = \int w(t)\, t^2 e^{-t^2} \dd{t},
    \qquad
    c_1 = \frac14 \int w(t)\, t^4 e^{-t^2} \dd{t}
}
depend only on the weight $w$.
By Taylor's theorem with the Lagrange form of the remainder, and since $1 + \e_a \geq 0$, the squared bias and sampling term each differ from their expansions above by at most $\abr{\e_a}^3 t^6/6$.
So the remainder in \cref{eq:app:slice_expansion} is at most $(\bb + 1) \abr{\e_a}^3 \int w(t)\, t^6 \dd{t} / 6$ in absolute value.

\begin{wrapfigure}{r}{0.35\linewidth}
    \centering
    \vspace{-1.5\baselineskip}
    \includegraphics[width=\linewidth]{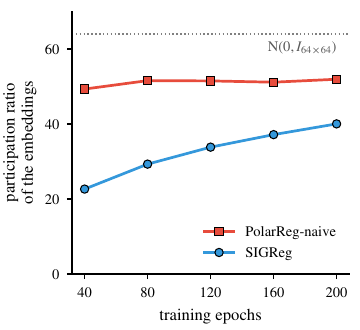}
    \caption{\textbf{Encoders trained with SIGReg keep a low participation ratio.}
    Participation ratio of the 64-dimensional embeddings of the Tiny-ImageNet-200 validation images during training. PR of SIGReg rises from 22.6 at epoch 40 to 40.0 at epoch 200.
    PR of PolarReg-naive is 49--53 from epoch 40.
    This shows that embeddings obtained with SIGReg are quite anisotropic in practice.
    }
    \label{fig:app:pr}
    \vspace{-1\baselineskip}
\end{wrapfigure}
\paragraph{Averaging over slices cancels the term linear in $\e_a$, and the result is independent of $\kk$.}
Each slice $a_k$ is uniform on $S^{d-1}$ and independent of the batch, so $\E\sbr{T(a_k)} = \E_a\sbr{\E\sbr{T(a) \mid a}}$ for every $k$, and the mean $T_\kk$ of the $\kk$ values has this expectation for every $\kk$.
The linear term vanishes because $\E\sbr{\e_a} = 0$, and \cref{eq:app:eps_moments} gives
\beq{
    \E\sbr{T_\kk} = c_0 + c_1 (\bb - 2)\, \frac{2 \omega^2}{d(d+2)} + O\rbr{(\bb + 1)\, \E\sbr{\abr{\e_a}^3}}.
    \label{eq:app:sliced_power}
}
This concludes our proof of \cref{prop:sliced_power}. \qed

\medskip

\noindent\refstepcounter{remark}\label{rem:app:pr}\textbf{Remark~\theremark} (Expectation of SIGReg and the participation ratio)\textbf{.}
In \cref{prop:sliced_power}, the expected value of SIGReg depends on the covariance $\Sigma$ only through $\omega$, its distance from identity.
Without the restriction to equal eigenvalues, $\operatorname{tr}(\Sigma^2) = d + \omega^2$, so $\omega^2/d^2 = 1/\operatorname{PR} - 1/d$, where
\beq{
    \operatorname{PR} = \frac{(\operatorname{tr}\Sigma)^2}{\operatorname{tr}(\Sigma^2)}
    \label{eq:app:pr}
}
is the participation ratio of $\Sigma$.
Two covariances with the same PR therefore have the same expected SIGReg value in \cref{eq:sliced_power}.
PR is the effective number of directions the variance spreads over, that is the effective dimensionality.
Each direction is accounted by the squared share of variance along it.
If $\Sigma$ has $r$ equal nonzero eigenvalues, then $\operatorname{PR} = r$.
If some directions carry more variance than others, $\operatorname{PR}$ is smaller.
$\text{N}(0, I_{64 \times 64})$, the target of SIGReg, has $\operatorname{PR} = 64$.
Sampling alone lowers it.
The covariance of 256 samples of $\text{N}(0, I_{64 \times 64})$ has $\operatorname{PR} \approx 51$.
A collapsed batch, with $\operatorname{PR} = 48$, is close to a control batch of the same size.
The two standard deviations in \cref{sec:blindness} make the same comparison along one slice.
Both equal $\sqrt{2(d/\operatorname{PR} - 1)/(d + 2)}$, that results in 0.10 at $\operatorname{PR} = 48$ and 0.09 at $\operatorname{PR} = 51$.
The same blindness appears in training.
On Tiny-ImageNet-200 with $d = 64$, the embeddings of the 10,000 validation images from an encoder trained with SIGReg have $\operatorname{PR} = 22.6$ at epoch 40 (\cref{fig:app:pr}).
PR only rises to $\sim$40.0 at epoch 200, showing that embeddings obtained with SIGReg are quite anisotropic in practice.
Consequently, this is also the range in which the test of \cref{sec:blindness} finds SIGReg weak. For PolarReg-naive, described in \cref{sec:decomposition}, PR is 49--53 from epoch 40.

\begin{remark}[Expectation of SIGReg on duplicated embeddings]
\label{rem:app:duplicates}
Let a batch of $\bb$ embeddings consist of $(1 - p)\bb$ independent samples of $\text{N}(0, I_{d \times d})$, of which $p\bb$ appear twice.
Then for any slices $a_1, \dots, a_\kk$, $\E\sbr{T_\kk} = (1 + 2p)\, c_0$, where $c_0$ is the same constant as in \cref{eq:app:sliced_power}, the expected value of $T_\kk$ on control batches.
\end{remark}
\begin{proof}
Along a slice $a$, projections of distinct samples are independent samples of $\text{N}(0, 1)$, with characteristic function $e^{-t^2/2}$, and empirical characteristic function $\varphi_a$ from \cref{eq:app:slice_stat}.
For each embedding, let $z_i(t) = e^{\i t\, a^\top u_i} - e^{-t^2/2}$, so that $\varphi_a(t) - e^{-t^2/2} = \frac{1}{\bb} \sum_{i=1}^{\bb} z_i(t)$.
If $u_i$ and $u_j$ are distinct samples, $z_i(t)$ and $z_j(t)$ are independent, both have mean zero, and each has $\E\sbr{\abr{z_i(t)}^2} = 1 - e^{-t^2}$ by \cref{eq:app:ecf} with $\bb = 1$.
A duplicated sample creates two equal $z_i(t)$, whose sum $2 z_i(t)$ has $\E\sbr{\abr{2 z_i(t)}^2} = 4\rbr{1 - e^{-t^2}}$; two independent samples lead to $2\rbr{1 - e^{-t^2}}$, since the expected cross term is zero.
So, with $(1 - 2p)\bb$ samples that appear once and $p\bb$ that appear twice,
\beqs{
    \E\sbr{\abr{\varphi_a(t) - e^{-t^2/2}}^2} = \frac{(1 - 2p)\bb + 4p\bb}{\bb^2}\rbr{1 - e^{-t^2}} = \rbr{1 + 2p} \frac{1 - e^{-t^2}}{\bb}.
}
Integrating against $\bb\, w(t)$, we have, $\E\sbr{T(a)} = (1 + 2p)\, c_0$ for every slice. Hence, $\E\sbr{T_\kk} = (1 + 2p)\, c_0$.
\end{proof}
At $\bb = 256$, one duplicated pair, $p = 1/256$, raises $\E\sbr{T_\kk}$ by 0.8\%, well inside its variation across control batches. Neither a larger batch nor more slices change the ratio.

\subsection{Polar decomposition of the isotropic Gaussian}
\label{app:polar}
\Cref{thm:polar} shows that a simple polar decomposition of the isotropic Gaussian is possible and exact. Here, we provide a short proof.

\begin{theorem}[Polar decomposition of an isotropic Gaussian]
\label{thm:app:polar}
Let $u$ be a nonzero random vector in $\reals^d$, and write $u = \rho\, \hat u$ with magnitude $\rho = \norm{u}$ and direction $\hat u = u/\norm{u}$ on the unit sphere $S^{d-1}$.
Then $u \sim \text{N}(0, I_{d \times d})$ if and only if $\rho \sim \chi_d$, $\hat u \sim \mathrm{Unif}(S^{d-1})$, and $\rho$ and $\hat u$ are independent.
\end{theorem}
\begin{proof}
This is a quite standard result on spherical distributions~\cite[Section 1.5]{muirhead1982aspects}. We give a short proof.
Suppose that $u \sim \text{N}(0, I_{d \times d})$.
Its $d$ coordinates are independent samples of $\text{N}(0, 1)$, so $\rho^2 = \norm{u}^2$, the sum of their squares, follows a chi-squared distribution with $d$ degrees of freedom, denoted by $\chi_d^2$.
For every orthogonal $d \times d$ matrix $H$, $Hu \sim \text{N}(0, I_{d \times d})$, so $Hu$ has the same distribution as $u$.
If a nonzero random vector has this property, its direction is uniform on $S^{d-1}$ and independent of its magnitude~\cite[Theorem 1.5.6]{muirhead1982aspects}.
This directly shows $\hat u \sim \mathrm{Unif}(S^{d-1})$, and $\rho$ and $\hat u$ to be independent.

Conversely, suppose that $\rho \sim \chi_d$, $\hat u \sim \mathrm{Unif}(S^{d-1})$, and $\rho$ and $\hat u$ are independent.
Let $g \sim \text{N}(0, I_{d \times d})$.
With the forward direction, $\norm{g} \sim \chi_d$ and $g/\norm{g} \sim \mathrm{Unif}(S^{d-1})$, and the two are independent.
The joint distribution of two independent variables is the product of their distributions, so the pairs $(\rho, \hat u)$ and $(\norm{g}, g/\norm{g})$ have the same joint distribution.
The vectors $u = \rho\, \hat u$ and $g = \norm{g}\, (g/\norm{g})$ are the same function of these two pairs, so they have the same distribution, and $u \sim \text{N}(0, I_{d \times d})$.
\end{proof}

\subsection{Distribution of inner products of uniform directions}
\label{app:cosine}
\Cref{thm:app:cosine} derives $\cosd$, the distribution of the inner product of two independent uniform directions; the target of the angular penalty in \cref{eq:angular}.
Quantiles of $\cosd$ are those of a Beta distribution, scaled and shifted to $[-1, 1]$.

\begin{theorem}[Distribution of inner products of uniform directions]
\label{thm:app:cosine}
Let $d \geq 2$, and let $\hat u$ and $\hat u'$ be independent and uniform on $S^{d-1}$.
Their inner product $s = \langle \hat u, \hat u' \rangle$ has the density
\beq{
    \cosd(s) = \frac{\Gamma(d/2)}{\sqrt\pi\, \Gamma((d-1)/2)} \rbr{1 - s^2}^{(d-3)/2}, \qquad s \in [-1, 1],
    \label{eq:app:cosine_density}
}
so $(s + 1)/2 \sim \text{Beta}(\b, \b)$ with $\b = (d-1)/2$.
The mean and variance of $s$ are 0 and $1/d$, respectively.
Conversely, let $\hat u$ and $\hat u'$ be two independent draws from some distribution on $S^{d-1}$. If their inner product has the density $\cosd$, then $\hat u$ is uniform on $S^{d-1}$~\cite[Proposition 1]{jiang2025hyperspheres}.
\end{theorem}
\begin{proof}
This is a standard result on spherical distributions~\cite[Section 1.5]{muirhead1982aspects}. We give a short derivation of the first part; the converse is from~\cite{jiang2025hyperspheres}.

\noindent
\textbf{The squared inner product follows a Beta distribution.}
Let $e$ be a unit vector.
The distribution of $\hat u$ does not change under orthogonal transformations, and $e e^\top$ is symmetric and idempotent with rank one.
So $\langle \hat u, e \rangle^2 = \hat u^\top e e^\top \hat u / \norm{\hat u}^2 \sim \text{Beta}(1/2, (d-1)/2)$~\cite[Theorem 1.5.7(ii)]{muirhead1982aspects}.
Since $\hat u$ is independent of $\hat u'$, it is still uniform given $\hat u'$.
So $s^2 = \langle \hat u, \hat u' \rangle^2 \sim \text{Beta}(1/2, (d-1)/2)$ as well.

\noindent
\textbf{The inner product is symmetric about zero, leading to its density.}
The vector $-\hat u$ is also uniform on $S^{d-1}$ and independent of $\hat u'$, so $s$ and $-s$ have the same distribution.
For $0 < x < 1$, the event $s^2 > x^2$ is a disjoint union of $s > x$ and $s < -x$. These have the same probability, so
\beqs{
    \P(s > x) = \frac12\, \P(s^2 > x^2) = \frac12 \int_{x^2}^1 h(t) \dd{t},
}
where $h(t) = \frac{\Gamma(d/2)}{\Gamma(1/2)\, \Gamma((d-1)/2)}\, t^{-1/2} (1 - t)^{(d-3)/2}$.
Differentiating in $x$ we get the density of $s$ at $x$, and by symmetry also at $-x$,
\beqs{
    -\dv{x} \P(s > x) = x\, h(x^2) = \frac{\Gamma(d/2)}{\sqrt\pi\, \Gamma((d-1)/2)} \rbr{1 - x^2}^{(d-3)/2} = \cosd(x),
}
since $\Gamma(1/2) = \sqrt\pi$.
Changing variables to $(s + 1)/2$ in $\cosd$ leads to a density $\text{Beta}(\b, \b)$.
Mean of $s$ is 0 by symmetry, and its variance $\E\sbr{s^2}$ is the mean of $\text{Beta}(1/2, (d-1)/2)$, which is $1/d$.
\end{proof}

The angular penalty compares the whole distribution of inner products with $\cosd$.
A rank collapse changes the second moment of this distribution. \Cref{rem:app:second_moment} computes this change, the variance of its finite-sample estimate on uniform directions, and a lower bound for batches constrained to $r$ dimensions.
Its last statement turns this second moment into a lower bound on the angular penalty itself.

\begin{remark}[Second moment of the distribution of inner products of uniform directions]
\label{rem:app:second_moment}
Let $\hat u_1, \dots, \hat u_\bb$ be unit vectors in $\reals^d$, and let $\overline{s^2}$ be the average of $s_{ij}^2 = \langle \hat u_i, \hat u_j \rangle^2$ over the $\bb(\bb-1)/2$ pairs.
(i) If the $\hat u_i$ are independent and uniform on $S^{d-1}$, $\overline{s^2}$ has mean $1/d$ and standard deviation about $2/(d\bb)$.
(ii) If the $\hat u_i$ lie in an $r$-dimensional subspace, $\overline{s^2} \geq (\bb/r - 1)/(\bb - 1)$.
(iii) For any $\hat u_i$, the angular penalty in \cref{eq:angular} on their inner products satisfies $\ell_{\hat u} \geq \big(\sqrt{\overline{s^2}} - \sqrt{\overline{q^2}}\big)^2$, where $\overline{q^2} \approx 1/d$ is the average of the squared targets in \cref{eq:angular}.
\end{remark}
\begin{proof}
\noindent
\textbf{For two random directions, the mean squared inner product is the inverse participation ratio.}
Let $\hat u$ and $\hat u'$ be independent draws from any distribution on $S^{d-1}$, with second-moment matrix $M = \E\sbr{\hat u \hat u^\top}$, and let $s = \langle \hat u, \hat u' \rangle$.
The scalar $s^2 = \hat u^\top \hat u' \hat u'^\top \hat u$ equals its trace, $\operatorname{tr}(\hat u \hat u^\top \hat u' \hat u'^\top)$.
Since $\hat u$ and $\hat u'$ are independent with the same distribution, $\E\sbr{s^2} = \operatorname{tr}(M^2)$.
Since $\operatorname{tr} M = \E\sbr{\norm{\hat u}^2} = 1$, the participation ratio of $M$ is $(\operatorname{tr} M)^2/\operatorname{tr}(M^2) = 1/\operatorname{tr}(M^2)$, and
\beqs{
    \norm{M - I/d}_F^2 = \operatorname{tr}(M^2) - \frac{2}{d}\operatorname{tr} M + \frac{1}{d} = \E\sbr{s^2} - \frac1d,
}
which is zero only for $M = I/d$.
Directions uniform on $r$ of the $d$ dimensions have $M = P_r/r$, where $P_r$ is the orthogonal projection onto them, so $\E\sbr{s^2} = \operatorname{tr}(M^2) = 1/r$.

\noindent
\textbf{(i) On uniform directions, the average $\overline{s^2}$ has mean $1/d$ and standard deviation about $2/(d\bb)$.}
Using the proof of \cref{thm:app:cosine}, $s_{ij}^2 \sim \text{Beta}(1/2, (d-1)/2)$, with mean $1/d$ and variance $2(d-1)/(d^2(d+2))$.
So $\overline{s^2}$ has mean $1/d$.
Two pairs that share one direction, say $(i, j)$ and $(i, l)$, are uncorrelated. Given $\hat u_i$, the terms $s_{ij}^2$ and $s_{il}^2$ are independent, and each has the mean $\hat u_i^\top \E\sbr{\hat u_j \hat u_j^\top} \hat u_i = 1/d$ whatever $\hat u_i$ is.
Two pairs that share no direction are independent.
Hence, the variance of $\overline{s^2}$ is the variance of one term divided by the number of pairs,
\beqs{
    \frac{4(d-1)}{d^2 (d+2)\, \bb (\bb-1)} \approx \rbr{\frac{2}{d\bb}}^2.
}

\noindent
\textbf{(ii) Constraining unit vectors to $r$ dimensions lower bounds their mean squared inner product.}
For $\bb$ unit vectors in an $r$-dimensional space, the sum of their squared inner products over all ordered pairs, including each vector with itself, is at least $\bb^2/r$~\cite[p.~397]{welch1974lower}.
Removing the $\bb$ terms of each vector with itself and averaging over the $\bb(\bb-1)$ ordered pairs gives
\beqs{
    \overline{s^2} \geq \frac{\bb^2/r - \bb}{\bb(\bb-1)} = \frac{\bb/r - 1}{\bb - 1}.
}
This is at most $1/d$ only when $\bb \leq (d - 1) r/(d - r)$.

\noindent
\textbf{(iii) The mean squared inner product bounds the angular penalty from below.}
Let $n = \bb(\bb-1)/2$, and for $x \in \reals^n$ let $\norm{x}_n = \norm{x}/\sqrt{n}$, the root mean square of its entries, which is a norm.
Let $\tilde s \in \reals^n$ hold the inner products in increasing order and $q \in \reals^n$ the targets of \cref{eq:angular}, so that $\ell_{\hat u} = \norm{\tilde s - q}_n^2$.
Sorting does not change a sum of squares, so $\norm{\tilde s}_n^2 = \overline{s^2}$, and $\norm{q}_n^2 = \overline{q^2}$ by definition.
With the triangle inequality we have $\norm{\tilde s - q}_n \geq \abr{\norm{\tilde s}_n - \norm{q}_n}$, and squaring both sides leads to the bound.
Since the targets are quantiles of $\cosd$, $\overline{q^2}$ is close to the second moment of $\cosd$, which is $1/d$ by \cref{thm:app:cosine}.
\end{proof}

To summarize, the angular penalty detects a rank collapse, however the directions are arranged.
If $\bb > (d - 1) r/(d - r)$, a batch confined to $r$ dimensions has a mean squared inner product larger than $1/d$, its mean on control batches.
So the angular penalty of such a batch has a positive lower bound.
On control batches, the angular penalty instead goes to zero as $1/\bb^2$ (\cref{sec:autodose}).
This is why the angular penalty rejects every collapsed batch in \cref{fig:rank_blindness}, even those optimized to look uniform.

\subsection{Properties of coupling embedding magnitude and direction by the alignment term}
\label{app:rearrangement}
\Cref{prop:rearrangement} shows that the alignment term is minimized only when images with larger view disagreement get smaller magnitude embeddings.
The view disagreement varies with the embedding direction, for instance with its cosine similarity to the class center, and hence so does the magnitude.
The independence of magnitude and direction in an isotropic Gaussian forbids this dependence, and works against the alignment term.
We also verify that the assumption of the proposition, that all views of an image have the same magnitude, holds closely on trained encoders.
Replacing the magnitudes of the views by their mean changes the alignment term of most images by less than 5\%.
In \cref{rem:app:coupling}, we prove that forgoing the independence term cannot result in collapsed embeddings.

\begin{proposition}
\label{prop:rearrangement}
Suppose the views of each image $i$ share a magnitude $\rho_i$, so their embeddings are $u_i^v = \rho_i\, \hat u_i^v$ with unit directions $\hat u_i^v$.
Let $\bar u_i = \vv^{-1} \sum_{v=1}^{\vv} \hat u_i^v$ be the mean direction of image $i$, and $D_i = \sum_{v=1}^{\vv} \norm{\hat u_i^v - \bar u_i}^2$ its view disagreement.
Then the alignment term of \cref{eq:lejepa} is
\beq{
    \frac{1}{\bb \vv d} \sum_{i=1}^{\bb} \sum_{v=1}^{\vv} \norm{u_i^v - m_i}^2
    = \frac{1}{\bb \vv d} \sum_{i=1}^{\bb} \rho_i^2\, D_i.
    \label{eq:rearrangement}
}
Swapping the magnitudes of two images does not change the magnitude regularizer $\ell_\r$ in \cref{eq:radial}.
Over all such swaps, the alignment term is smallest only when images whose views disagree more have smaller magnitudes: $D_i < D_j$ implies $\rho_i \geq \rho_j$.
If $D_i$ depends on the direction of the image embedding, then so does $\rho_i$, so $\rho$ and $\hat u$ are not independent.
\end{proposition}

\paragraph{Proof of \cref{prop:rearrangement}.}
With a shared magnitude of embeddings for all views of an image, the alignment term weighs view disagreement of each image by its squared magnitude.
The mean of views of image $i$ is $m_i = \rho_i\, \bar u_i$, so $u_i^v - m_i = \rho_i \rbr{\hat u_i^v - \bar u_i}$.
Squaring, summing over views and images, and dividing by $\bb \vv d$ results in \cref{eq:rearrangement}.

\paragraph{Swapping magnitudes of two images changes neither the magnitude regularizer nor the view disagreements.}
The magnitude regularizer compares the distribution of magnitudes in the batch with $\chi_d$, and a swap does not alter it.
Of course, a swap keeps every $D_i$, unchanged.
Sequences of swaps reach every rearrangement of magnitudes among images, so a minimum of the alignment term over swaps maximizes $\sum_i (-D_i)\, \rho_i^2$ over all rearrangements of the squared magnitudes.

\paragraph{To minimize the alignment term, images whose views disagree more must have smaller magnitudes.}
For real numbers $a_1, \dots, a_\bb$ and $b_1, \dots, b_\bb$, if $\sum_i a_i b_i$ is at least as large as for every rearrangement of $b_i$, then $(a_i - a_j)(b_i - b_j) \geq 0$ for all $i$ and $j$~\cite[Theorem 369]{hardy1934inequalities}.
With $a_i = -D_i$ and $b_i = \rho_i^2$, we have $(D_j - D_i)(\rho_i^2 - \rho_j^2) \geq 0$. Since magnitudes are nonnegative, $D_i < D_j$ implies $\rho_i \geq \rho_j$.
Hence, to minimize the alignment term, the magnitude of an image must be a strictly decreasing function of its view disagreement, $\rho_i = h(D_i)$.
Since $h$ is one-to-one, two regions of the sphere whose images differ in view disagreement also differ in magnitude.
To conclude, in this scenario, $\rho$ and $\hat u$ are not independent. \qed

\begin{wrapfigure}{r}{0.35\linewidth}
    \centering
    \includegraphics[width=\linewidth]{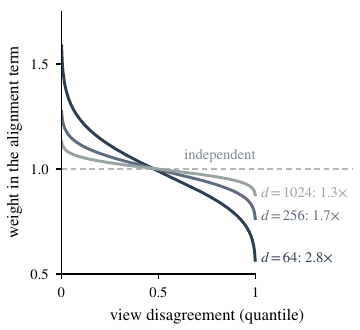}
    \caption{\textbf{Without independence, ambiguous images get less weight in the alignment term.}
    Weight $\rho_i^2/d$ of each image in the alignment term against its $D_i$ quantile, for a batch of 256 images with exact $\chi_d$ magnitudes.
    Under independence the weight is flat; otherwise it falls with view disagreement.
    At the minimum of the alignment term, the image at $D_i$ quantile $x$ gets the $(1 - x)$-quantile of $\chi_d$ as its magnitude.}
    \label{fig:app:weight}
    \vspace{-3.5\baselineskip}
\end{wrapfigure}

\Cref{fig:app:weight} shows the resulting weight $\rho_i^2/d$ of each image in the alignment term, at $d = 64, 256$ and $1024$. However, \cref{prop:rearrangement} assumes that all view embeddings of an image share a magnitude. The following discussion argues why this assumption is mild.

\paragraph{On trained encoders, shared magnitude of image view embeddings is not a strong assumption.}
The magnitude regularizer compares the magnitudes of all $\bb\vv$ embeddings in the batch with $\chi_d$. It does not force the views of an image to share a magnitude.
Write the magnitude of image $i$ and view $v$ as $\bar\rho_i + \e_i^v$, where $\bar\rho_i$ is the mean magnitude over views, and $\sum_{v=1}^{\vv} \e_i^v = 0$. Let $\mu_i = \vv^{-1} \sum_{v=1}^{\vv} \e_i^v\, \hat u_i^v$.
Then $u_i^v - m_i = \bar\rho_i \rbr{\hat u_i^v - \bar u_i} + \rbr{\e_i^v\, \hat u_i^v - \mu_i}$.
Summing the squares over views, we have
\beqs{
    \aed{
    \sum_{v=1}^{\vv} \norm{u_i^v - m_i}^2
    = &\bar\rho_i^2 D_i
    + \sum_{v=1}^{\vv} \norm{\e_i^v\, \hat u_i^v - \mu_i}^2 \\
    &- 2 \bar\rho_i \sum_{v=1}^{\vv} \e_i^v \inner{\bar u_i}{\hat u_i^v - \bar u_i}.
    }
}
The first term is the one in \cref{eq:rearrangement}, with $\rho_i = \bar\rho_i$.
The second is at most $\sum_{v=1}^{\vv} (\e_i^v)^2$, that is $\vv$ times the variance of magnitudes of the views.
By Cauchy--Schwarz inequality, the third is at most $2 \bar\rho_i \sqrt{D_i \sum_{v=1}^{\vv} (\e_i^v)^2}$ in absolute value.
Both vanish when the views share a magnitude.
Relative to the first term,
both are small when standard deviation of magnitudes of the views is much smaller than $\bar\rho_i \sqrt{D_i / \vv}$.
For PolarReg-naive trained on Tiny-ImageNet-200, with 8 augmented views of the 10,000 validation images, the alignment term is within 5\% of the first term for 91\% of images at $d = 64$ and for 97\% at $d = 256$.
If one sums the alignment term over all images, as in the loss, the first term matches it within 0.5\% for PolarReg-naive and within 3\% for SIGReg, at both embedding dimensions.

\Cref{prop:rearrangement} suggests that enforcing independence between magnitude and direction works against the alignment term.
However, without independence, magnitude can depend on direction freely, as long as the magnitude follows $\chi_d$ and the direction is uniform.
\Cref{rem:app:coupling} shows that this freedom, provably does not lead to any feature collapse.

\begin{remark}[Moments of embeddings without independence]
\label{rem:app:coupling}
Let $\rho \sim \chi_d$ and $\hat u \sim \mathrm{Unif}(S^{d-1})$ have any joint distribution, and let embedding $u = \rho\, \hat u \in \reals^d$.
Then $\E\sbr{\norm{u}^2} = d$, and for every unit vector $e \in S^{d-1}$,
\beqs{
    \abs{e^\top \E\sbr{u u^\top} e - 1} \leq c_d = 2 \sqrt{\frac{d-1}{d(d+2)}} < \frac{2}{\sqrt d},
    \qquad
    \abs{\inner{e}{\E\sbr{u}}} \leq \frac{\operatorname{sd}(\rho)}{\sqrt d}.
}
So every eigenvalue of $\E\sbr{u u^\top}$ lies in $[1 - c_d, 1 + c_d]$, and $\norm{\E\sbr{u}} \leq \operatorname{sd}(\rho)/\sqrt d$, where $\operatorname{sd}(\rho)$ is the standard deviation of $\rho$.
To summarize, even without independence, eigenvalues of the second-moment matrix of the embeddings are close to 1, and the mean of the embeddings stays close to the origin.
\end{remark}
\begin{proof}
\noindent
Let $c = \inner{e}{\hat u}$.
By \cref{thm:app:cosine}, $c^2 \sim \text{Beta}(1/2, (d-1)/2)$, with mean $1/d$ and variance $2(d-1)/(d^2(d+2))$.
Since $\rho \sim \chi_d$, $\rho^2$ is distributed with a mean $d$ and variance $2d$.
So $\E\sbr{\norm{u}^2} = \E\sbr{\rho^2} = d$, and
\beqs{
    e^\top \E\sbr{u u^\top} e = \E\sbr{\inner{e}{u}^2} = \E\sbr{\rho^2 c^2} = \E\sbr{\rho^2}\, \E\sbr{c^2} + \cov\rbr{\rho^2, c^2} = 1 + \cov\rbr{\rho^2, c^2}.
}

\noindent
\textbf{For every joint distribution, eigenvalues of the second-moment matrix stay close to 1.}
In absolute value, the covariance is at most product of the standard deviations, $c_d = \sqrt{2d} \cdot \sqrt{2(d-1)/(d^2(d+2))}$, and $c_d < 2/\sqrt d$ since $d - 1 < d + 2$.
Eigenvalues of $\E\sbr{uu^\top}$ lie between the smallest and largest value of $e^\top \E\sbr{uu^\top} e$ over unit vectors $e \in S^{d-1}$. Therefore the eigenvalues are in $[1 - c_d, 1 + c_d]$.

\noindent
\textbf{The mean of the embeddings stays close to origin.}
Since $-\hat u$ is also uniform, $\E\sbr{c} = 0$, and the variance of $c$ is $\E\sbr{c^2} = 1/d$.
Hence $\inner{e}{\E\sbr{u}} = \E\sbr{\rho c} = \cov(\rho, c)$, which by Cauchy--Schwarz inequality is at most $\operatorname{sd}(\rho)/\sqrt d$ in absolute value.
Taking $e$ along $\E\sbr{u}$ bounds its norm.
\end{proof}

For perspective, at $d = 256$, $c_d = 0.12$ and $\operatorname{sd}(\rho)/\sqrt d = 0.044$: every eigenvalue lies between 0.88 and 1.12, and the mean is within 0.044 of the origin.
So no joint distribution with these marginals collapses the embeddings: their mean squared magnitude stays at $d$, and every eigenvalue of $\E\sbr{u u^\top}$ is at least $1 - c_d$, so there is no rank collapse.
Without independence, distribution of the magnitudes can couple with the directions in interesting ways, which it cannot for the isotropic Gaussian.
For example, magnitude can peak in some directions, say class centroids and dip at class boundaries, as in \cref{fig:shape}a,b. Magnitude still follows $\chi_d$ and the direction stays uniform; trained encoders consistently have this shape (see \cref{fig:shape}c, d).

\subsection{Details of PolarReg}
\label{app:embedding_distance}
\begin{figure}[t]
    \centering
    \includegraphics[width=0.9\linewidth]{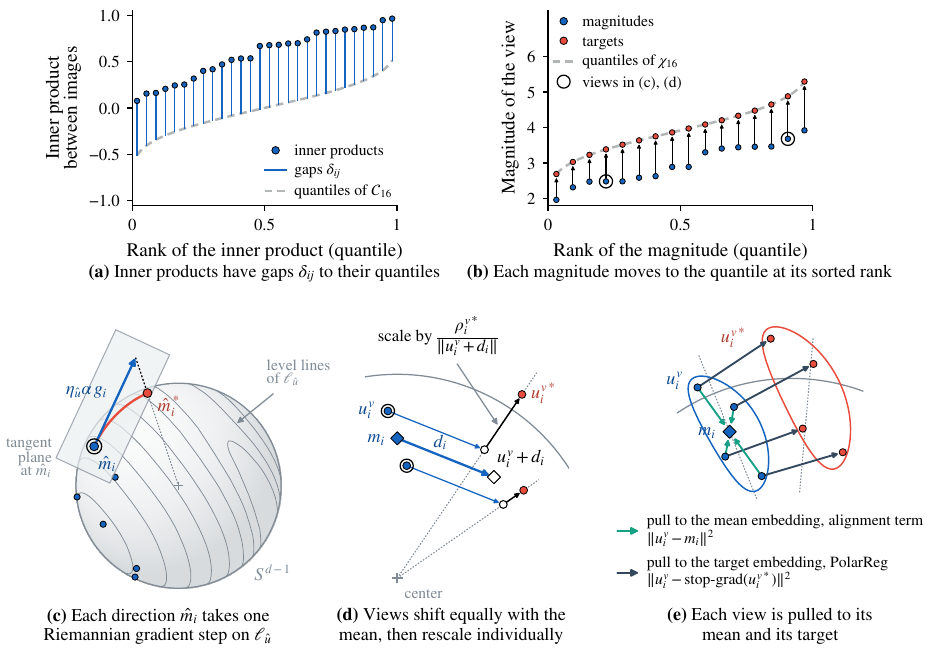}
    \caption{\textbf{Illustration of the construction of PolarReg in \cref{eq:ot_loss} from the magnitude and angular regularizers in \cref{eq:radial,eq:angular}.}
    The targets of \cref{eq:ot_loss} on a toy batch in three dimensions.
    The batches are shown in blue, the targets are shown in red and the references if any are shown in gray. 
    \textbf{(a)} Inner products between image directions, sorted, against quantiles of $\cosd[16]$, with gaps $\d_{ij}$.
    \textbf{(b)} Magnitudes of embeddings, sorted, against quantiles of $\chi_{16}$, each with an arrow to the quantile at its sorted rank.
    \textbf{(c)} The field $g_i$ lies in the tangent plane at $\hat m_i$, along the descent direction of $\ell_{\hat u}$.
    The step $\eta_{\hat u} \a\, g_i$ leaves the sphere, and normalization brings it back along the ray to the center, which gives the target direction $\hat m_i^*$. The drawing is inspired by \citet[Chapter~3]{boumal2023introduction}.
    \textbf{(d)} The mean embedding $m_i$ of the views changes its direction to $\hat m_i^*$, without changing its magnitude. It moves by $d_i = \norm{m_i} (\hat m_i^* - \hat m_i)$. Every view $u_i^v$ shifts by the same $d_i$, to $u_i^v + d_i$ (hollow). It is then rescaled along its ray from the center by $\rho_i^{v*} / \norm{u_i^v + d_i}$, to its target magnitude from (b). This gives us the target embedding $u_i^{v*}$.
    \textbf{(e)} A schematic of \cref{eq:ot_loss} for one image: each view is pulled toward the mean embedding $m_i$ by the alignment term and toward its target $u_i^{v*}$ by PolarReg. This formulation replaces both regularizers of PolarReg-naive.
}
    \label{fig:transport}
    \vspace{-0.5\baselineskip}
\end{figure}

The optimal coefficients $\l_\r$ and $\l_{\hat u}$ of PolarReg-naive depend on the batch size $\bb$, embedding dimension $d$, and the structure in the data (\cref{sec:autodose}).
For instance, the angular penalty averages over $\bb(\bb-1)/2$ pairs and ignores the magnitudes of the embeddings, so its gradient with respect to an embedding falls with $\bb$ and as $1/\norm{m_i}$, while the gradient of the alignment term grows with the magnitudes.
One way to remove these dependencies is to precondition the gradient of each penalty with its inverse Hessian, as in a Gauss--Newton step.
Since both penalties here, are squared 2-Wasserstein distances, they are quadratic, and this Newton step is simple.
We write the result of this step as a target embedding $u_i^{v*}$, and pull $u_i^v$ towards it with $\Om_{\text{PolarReg}}$ in \cref{eq:ot_loss}, a squared distance between embeddings.
This introduces step sizes $\eta_\r$ and $\eta_{\hat u}$ which do not change with the batch size, embedding dimension or magnitudes of the embeddings (\cref{prop:ot_step}).
\Cref{fig:transport} illustrates the construction on a toy batch.

\subsubsection{Construction of the target embeddings for PolarReg}
\label{app:ot_construction}

\Cref{sec:autodose} builds the target $u_i^{v*}$ of each embedding in three steps. It moves the magnitudes (\cref{eq:moving_magnitude}), moves the directions of the mean embeddings (\cref{eq:moving_direction}), and combines the two into $u_i^{v*}$.
Here we provide more details of each step.
The first two steps use a property of \cref{eq:w2}.
For $n$ samples with quantiles $q_1 \leq \dots \leq q_n$, its gradient with respect to the $r$-th smallest sample $z_{(r)}$ is $\frac{2}{n}(z_{(r)} - q_r)$, and its Hessian is $\frac{2}{n} I$.
So a Newton step moves each sample to the quantile at its desired sorted rank.

\paragraph{For the magnitude regularizer, each target magnitude fixes a target embedding.}
The $r$-th smallest magnitude of the $\bb\vv$ views of a batch should equal the quantile $q_r$ of $\chi_d$.
The target magnitude $\rho_i^{v*} = \rho_i^v + \eta_\r \rbr{q_i^v - \rho_i^v}$ in \cref{eq:moving_magnitude} moves each magnitude to its quantile $q_i^v$ (\cref{fig:transport}b) with a step size $\eta_\r$.
With $\eta_\r > 1$, the target lies past the quantile, compensating for the alignment term pulling the magnitudes down.

\paragraph{For the angular regularizer, each direction affects $\bb - 1$ inner products.}
The $r$-th smallest of the $\bb(\bb-1)/2$ inner products $s_{ij}$ in \cref{eq:angular} should equal the quantile $q_r$ of $\cosd$.
Let $\d_{ij} = q_{ij} - s_{ij}$ be the gap (\cref{fig:transport}a) between $s_{ij}$ and its quantile $q_{ij}$.
The target of $s_{ij}$ is shared by images $i$ and $j$.
In general, no set of directions satisfies all the inner product targets.
When $\bb > 2d - 1$, the inner products outnumber the $\bb(d-1)$ free coordinates of the directions.
Inner products assigned by the sorted quantiles may not be realizable either.
\Cref{eq:moving_direction} therefore moves each direction by a step that best satisfies the targets, $\hat m_i^* \propto \hat m_i + \eta_{\hat u} \D_i$, with $\D_i \perp \hat m_i$ tangent to the sphere.
To find the steps, consider the moved directions $\hat m_i' = (\hat m_i + \D_i)/\norm{\hat m_i + \D_i}$, with inner products $s_{ij}' = \inner{\hat m_i'}{\hat m_j'}$.
The best step $\D = (\D_1, \dots, \D_\bb)$ minimizes
\beq{
    \sum_{i < j} \rbr{s_{ij}' - q_{ij}}^2.
    \label{eq:app:ot_nls}
}
This is a nonlinear least-squares problem, since each $s_{ij}'$ is a nonlinear function of the steps $\D_i$ and $\D_j$.

\paragraph{Linearizing the inner products, we have a linear least-squares problem.}
Since $\D_i \perp \hat m_i$, $\norm{\hat m_i + \D_i} = (1 + \norm{\D_i}^2)^{1/2} = 1 + O(\norm{\D_i}^2)$, so the normalization changes the inner products $s_{ij}'$ only at second order.
The inner product $\inner{\hat m_i + \D_i}{\hat m_j + \D_j} = s_{ij} + \inner{\D_i}{\hat m_j} + \inner{\hat m_i}{\D_j} + \inner{\D_i}{\D_j}$ has one first-order term for each direction of the pair, and its last term is of second order.
With $q_{ij} = s_{ij} + \d_{ij}$,
\beqs{
    s_{ij}' - q_{ij} = \inner{\D_i}{\hat m_j} + \inner{\hat m_i}{\D_j} - \d_{ij} + O(\norm{\D}^2),
}
is the first-order Taylor expansion of $s_{ij}'$ at $\D = 0$.
Minimizing the sum of squares of the linear part over all steps is one Gauss--Newton step for \cref{eq:app:ot_nls}.
It is a linear least-squares problem with $\bb(\bb-1)/2$ equations and $\bb(d-1)$ unknowns.

\paragraph{We restrict the step to the descent direction of the angular regularizer.}
Solving this full linear problem at every training step is costly.
Instead, we fix each step to the descent direction of the angular regularizer on the sphere (\cref{prop:ot_step}a), that is $\D_i = t\, g_i$ with
\beqs{
    g_i = \rbr{I - \hat m_i \hat m_i^\top} \sum_{j \neq i} \d_{ij}\, \hat m_j.
}
This leaves a single unknown, the scale $t$, and the problem becomes a line search along $g = (g_1, \dots, g_\bb)$. With this, each linearized residual becomes $t\, c_{ij} - \d_{ij}$ with $c_{ij} = \inner{g_i}{\hat m_j} + \inner{\hat m_i}{g_j}$. The sum of their squares, $\sum_{i<j} (t\, c_{ij} - \d_{ij})^2$, is a parabola in $t$ that is smallest at $t = \inner{c}{\d}/\norm{c}^2$. Here $c = (c_{ij})_{i<j}$ and $\d = (\d_{ij})_{i<j}$ are vectors in $\reals^{\bb(\bb-1)/2}$, with one entry for each pair of images.
We call this minimizer $\a$.
So $\a$ is a Newton step along $g$, which is how it depends on the Hessian of \cref{eq:angular}.
Like $\eta_\r$ for the magnitudes, $\eta_{\hat u}$ scales this step, and the target direction in \cref{eq:moving_direction} is $\hat m_i^* \propto \hat m_i + \eta_{\hat u} \a\, g_i$ (\cref{fig:transport}c).
Normalization is a retraction, that brings the step back on the sphere and matches moving along it, to first order~\cite[Example~3.49]{boumal2023introduction}.
This operation is one step of Riemannian gradient descent on the directions~\cite[Algorithm~4.1]{boumal2023introduction}.

\paragraph{Combining the angular shift and the magnitude rescaling provides a target embedding for every view.}
Shifting every view of image $i$ by $d_i = \norm{m_i} \hat m_i^* - m_i$ turns the direction of their mean embedding to $\hat m_i^*$, without changing its magnitude $\norm{m_i}$. This transformation keeps the offset of each view from the mean (\cref{fig:transport}d).
Since $g_i \perp \hat m_i$, $\norm{\hat m_i + \eta_{\hat u} \a\, g_i} = 1 + O(\eta_{\hat u}^2 \a^2 \norm{g_i}^2)$. So $d_i = \eta_{\hat u} \a \norm{m_i}\, g_i$ to first order; a displacement in the embedding space.
Each view is then rescaled to its target magnitude, leading to the target embedding,
\beqs{
    u_i^{v*} = \r_i^{v*}\, \frac{u_i^v + d_i}{\norm{u_i^v + d_i}},
}
as shown in \cref{fig:transport}d.
Regressing onto these target embeddings gives $\Om_{\text{PolarReg}}$ in \cref{eq:ot_loss}, replacing $\Om_{\text{PolarReg-naive}}$. In total, each view $u_i^v$ is pulled toward its target $u_i^{v*}$ and its mean embedding $m_i$ (\cref{fig:transport}e).

\subsubsection{Properties of PolarReg}
\label{app:ot_step}
\Cref{prop:ot_step} compares $\Om_{\text{PolarReg}}$ in \cref{eq:ot_loss} with the magnitude and angular regularizers in \cref{eq:radial,eq:angular}.
To first order, $\Om_{\text{PolarReg}}$ pulls embeddings along gradients of the magnitude and angular regularizers, with coefficients set by the batch.
The gradient of $\Om_{\text{PolarReg}}$ is zero if and only if the gradients of both regularizers are zero.

\begin{proposition}
\label{prop:ot_step}
Let $\eta_\r, \eta_{\hat u} > 0$, and let $n = \bb(\bb-1)/2$ be the number of pairs of images.
Assume that the embeddings $u_i^v$ and the mean embeddings $m_i$ are nonzero, the inner products $s_{ij}$ are distinct, and the magnitudes of the embeddings are distinct. These assumptions ensure that $\ell_\r$ and $\ell_{\hat u}$ in \cref{eq:radial,eq:angular} are differentiable. The following hold.

\begin{enumerate}[(a), itemsep=1ex, topsep=0.0ex]
\item The descent direction $g_i$ is $-n/2$ times the Riemannian gradient of $\ell_{\hat u}$ at $\hat m_i$ on the sphere, and $\a > 0$ unless $g = (g_1, \dots, g_\bb) = 0$. Hence the target direction $\hat m_i^*$ moves every direction $\hat m_i$ along the step that gradient descent on \cref{eq:angular} takes.

\item The step $\D_i = \a\, g_i$ changes each inner product $s_{ij}$ by $\a\, c_{ij}$ to first order. The mean of the squared changes, $\norm{\a c}^2/n$, is at most $\ell_{\hat u}$, for any batch size and any scale of the embeddings.
In contrast, the gradient of $\ell_{\hat u}$ with respect to an embedding $u_i^v$ is inversely proportional to the number of pairs $n$ and magnitude $\norm{m_i}$.

\item To first order in $u_i^{v*} - u_i^v$, the gradient of $\Om_{\text{PolarReg}}$ with respect to $u_i^v$ is the gradient of the magnitude regularizer of $\Om_{\text{PolarReg-naive}}$ in \cref{eq:full_decomposition} with $\l_\r = \eta_\r/d$, plus the component orthogonal to $u_i^v$ of the gradient of its angular regularizer with $\l_{\hat u} = \eta_{\hat u} \a (\bb-1) \norm{m_i}^2/(2d)$.

\item The target embeddings equal the embeddings, $u_i^{v*} = u_i^v$ for all $v$ and $i$, if and only if every magnitude $\rho_i^v$ equals its quantile $q_i^v$ and the directions $\hat m_1, \dots, \hat m_\bb$ are a critical point of $\ell_{\hat u}$.
Hence the gradient of $\Om_{\text{PolarReg}}$ is zero if and only if the gradients of both regularizers in \cref{eq:radial,eq:angular} are zero.

\item For every view, $\norm{u_i^{v*} - u_i^v} < \eta_\r \lvert q_i^v - \rho_i^v \rvert + 2\sqrt2\, \norm{m_i}$.
Hence the gradient of $\Om_{\text{PolarReg}}$ is bounded by the magnitudes of the mean embeddings and the absolute differences between the magnitudes and their quantiles, irrespective of the scale $\a$ and gaps $\d_{ij}$.
\end{enumerate}
\end{proposition}

\begin{proof}
Let $P_i = I - \hat m_i \hat m_i^\top$ be the orthogonal projection onto vectors orthogonal to $\hat m_i$---the tangent space of the unit sphere at $\hat m_i$.
As discussed earlier, $\ell_{\hat u} = \frac1n \sum_{i<j} \rbr{s_{ij} - q_{ij}}^2 = \frac1n \sum_{i<j} \d_{ij}^2$.

\paragraph{Proof of (a).}
Treating $\hat m_i$ as a vector in $\reals^d$, $\partial s_{ij}/\partial \hat m_i = \hat m_j$, and
\beqs{
    \frac{\partial \ell_{\hat u}}{\partial \hat m_i} = \frac2n \sum_{j \neq i} \rbr{s_{ij} - q_{ij}}\, \hat m_j = -\frac2n \sum_{j \neq i} \d_{ij}\, \hat m_j.
}
A tangent vector $\xi \perp \hat m_i$ moves $\hat m_i$ along the curve $(\hat m_i + t\xi)/\norm{\hat m_i + t\xi}$.
The normalization in the denominator changes $\hat m_i + t\xi$ only to second order in $t$.
The curve has derivative $\xi$ at $t = 0$; $\ell_{\hat u}$ changes along the curve at the rate $\inner{\partial \ell_{\hat u}/\partial \hat m_i}{\xi} = \inner{P_i\, \partial \ell_{\hat u}/\partial \hat m_i}{\xi}$, since $\xi \perp \hat m_i$.
The Riemannian gradient of $\ell_{\hat u}$ at $\hat m_i$ is the tangent vector whose inner product with every tangent $\xi$ has this derivative~\cite[Definition~3.58]{boumal2023introduction}. Hence it is $P_i\, \partial \ell_{\hat u}/\partial \hat m_i = -(2/n)\, g_i$~\cite[Proposition~3.61]{boumal2023introduction}.

\noindent
Now let $a_i = \sum_{j \neq i} \d_{ij}\, \hat m_j$, so that $g_i = P_i a_i$.
Since $\d_{ij} = \d_{ji}$, we have
\beqs{
    \inner{c}{\d} = \sum_{i<j} \d_{ij} \rbr{\inner{g_i}{\hat m_j} + \inner{\hat m_i}{g_j}} = \sum_i \inner{g_i}{a_i} = \sum_i \inner{P_i a_i}{P_i a_i} = \sum_i \norm{g_i}^2,
}
where the third equality holds because $P_i$ is an orthogonal projection.
So $\a = \sum_i \norm{g_i}^2 / \norm{c}^2 > 0$ when $g \neq 0$. The target direction $\hat m_i^* \propto \hat m_i + \eta_{\hat u} \a\, g_i$ moves $\hat m_i$ along the great circle in the direction $g_i$.

\paragraph{Proof of (b).}
By \cref{app:ot_construction}, the step $t\, g_i$ changes $s_{ij}$ by $t\, c_{ij}$ to first order in $t$, and $\a$ minimizes $\sum_{i<j} (t\, c_{ij} - \d_{ij})^2 = t^2 \norm{c}^2 - 2t \inner{c}{\d} + \norm{\d}^2$.
Then $\a c$ is the orthogonal projection of $\d$ onto the line through $c$, so $\norm{\a c} \leq \norm{\d}$.
The entries of $\a c$ and $\d$ are the changes $\a\, c_{ij}$ and gaps $\d_{ij}$ of the $n$ pairs of images. The mean of the squared changes is $\norm{\a c}^2/n \leq \norm{\d}^2/n = \ell_{\hat u}$, the mean of the squared gaps.
Since the inner products depend only on the directions $\hat m_i$, this bound holds for any batch size, embedding dimension and scale of the embeddings.

\paragraph{Proof of (c).}
By \cref{app:ot_construction}, $d_i = \eta_{\hat u} \a \norm{m_i}\, g_i$ to first order in the step size $\eta_{\hat u} \a \norm{g_i}$, and $u_i^{v*} = \rho_i^{v*}\, (u_i^v + d_i) / \norm{u_i^v + d_i}$.
To first order in $\norm{d_i}$, $\norm{u_i^v + d_i} = \rho_i^v + \inner{\hat u_i^v}{d_i}$, so
\beqs{
    u_i^{v*} - u_i^v = \rbr{I - \hat u_i^v (\hat u_i^v)^\top} d_i + \eta_\r \rbr{q_i^v - \rho_i^v}\, \hat u_i^v.
}
The magnitude of the embedding is only affected by the target magnitude, and its direction only through the target direction of the mean embedding.
With the fixed target embeddings, the gradient of $\Om_{\text{PolarReg}}$ with respect to $u_i^v$ is $2 (u_i^v - u_i^{v*})/(\bb \vv d)$, which to first order (in the step size) is
\beqs{
    -\frac{2}{\bb \vv d} \sbr{\eta_{\hat u} \a \norm{m_i} \rbr{I - \hat u_i^v (\hat u_i^v)^\top} g_i + \eta_\r \rbr{q_i^v - \rho_i^v}\, \hat u_i^v}.
}
The gradients of $\Om_{\text{PolarReg-naive}}$ in \cref{eq:full_decomposition} with respect to $u_i^v$ are $-2 \l_\r (q_i^v - \rho_i^v)\, \hat u_i^v/(\bb \vv)$ for the magnitude regularizer and $-2 \l_{\hat u}\, g_i/(n \vv \norm{m_i})$ for the angular regularizer.
Comparing, the magnitude parts agree for $\l_\r = \eta_\r/d$, and the angular parts agree for $\l_{\hat u} = \eta_{\hat u} \a (\bb-1) \norm{m_i}^2/(2d)$, except that $\Om_{\text{PolarReg}}$ keeps only the component orthogonal to $u_i^v$.
In $\Om_{\text{PolarReg-naive}}$, the angular gradient $-2 \l_{\hat u}\, g_i/(n \vv \norm{m_i})$ is orthogonal to $\hat m_i$ but in general not to $u_i^v$. Its component along $\hat u_i^v$, proportional to $\langle g_i, \hat u_i^v \rangle$, changes the magnitude of each view, so a gradient step on the angular regularizer also moves the magnitudes, while in $\Om_{\text{PolarReg}}$ only the target magnitude does.

\paragraph{Proof of (d).}
If every magnitude equals its quantile and $g = 0$, then $d_i = 0$ and $\rho_i^{v*} = \rho_i^v$, hence $u_i^{v*} = u_i^v$.
Conversely, suppose that $u_i^{v*} = u_i^v$ for every embedding.
Their magnitudes are equal, so $\eta_\r (q_i^v - \rho_i^v) = \rho_i^{v*} - \rho_i^v = 0$. Every magnitude equals its quantile since $\eta_\r > 0$.
Substituting $\rho_i^{v*} = \rho_i^v$ into $u_i^v = u_i^{v*} = \rho_i^{v*}\, (u_i^v + d_i)/\norm{u_i^v + d_i}$ we have $u_i^v + d_i$ to be a positive multiple of $u_i^v$, for every view.
If $d_i \neq 0$, every view of image $i$, and so $m_i$, would be parallel to $d_i$.
But $d_i$ has the component $\eta_{\hat u} \a \norm{m_i}\, g_i / \norm{\hat m_i + \eta_{\hat u} \a\, g_i}$ orthogonal to $\hat m_i$, which is nonzero whenever $d_i \neq 0$.
Hence $d_i = 0$, which means $\hat m_i^* = \hat m_i$ and $\a g_i = 0$ for every $i$.
By (a), $\a > 0$ unless $g = 0$, so $g = 0$, and the Riemannian gradient of $\ell_{\hat u}$, $-(2/n)\, g_i$ at each $\hat m_i$, is zero.
The gradients of the two regularizers in \cref{eq:radial,eq:angular} are zero when every magnitude equals its quantile and $g = 0$, which proves (d).

\paragraph{Proof of (e).}
Note that $u_i^{v*} = \rho_i^{v*}\, e$ and $u_i^v + d_i = \norm{u_i^v + d_i}\, e$ with the unit vector $e = (u_i^v + d_i)/\norm{u_i^v + d_i}$. Hence,
\beqs{
    \norm{u_i^{v*} - u_i^v} \leq \lvert \rho_i^{v*} - \norm{u_i^v + d_i} \rvert + \norm{d_i} \leq \lvert \rho_i^{v*} - \rho_i^v \rvert + 2 \norm{d_i},
}
where the second inequality uses the reverse triangle inequality $\lvert \norm{u_i^v + d_i} - \rho_i^v \rvert \leq \norm{d_i}$.
The first term is $\eta_\r \lvert q_i^v - \rho_i^v \rvert$.
Since $g_i \perp \hat m_i$, the target direction $\hat m_i^*$ has an angle $\psi_i = \arctan(\eta_{\hat u} \a \norm{g_i}) < \pi/2$ with $\hat m_i$, so $\norm{d_i} = \norm{m_i} \norm{\hat m_i^* - \hat m_i} = 2 \norm{m_i} \sin(\psi_i/2) < \sqrt2\, \norm{m_i}$.
\end{proof}

This shows that $\Om_{\text{PolarReg}}$ enforces the same distributions of the magnitudes and directions as $\Om_{\text{PolarReg-naive}}$, with coefficients set by each batch.
Its gradient is bounded by the magnitudes of the embeddings and their quantiles, and it is zero only where the gradients of the regularizers in \cref{eq:radial,eq:angular} are zero.

\section{Experimental details}
\label{app:experiments}

\subsection{Implementation of the independence regularizer in \cref{eq:independence}}
\label{app:independence}

\Cref{sec:decomposition} enforces independence of magnitude and direction by penalizing $\text{MMD}^2$. Results with this enforcement are presented in the ``PolarReg-naive + Independence'' row of \cref{tab:independence}.
In this section, we provide details on which views $\ell_\perp$ uses, why it does not backpropagate through the magnitudes and bandwidths, its coefficient $\l_\perp$, and how we measure $\text{MMD}^2$ in \cref{tab:independence}.

In \cref{eq:independence}, we average only over the global views of each image; there are 2 global views on ImageNet-100.
A constant magnitude makes $\text{MMD}^2$ zero for any set of directions. Hence, a gradient of $\ell_\perp$ with respect to the magnitudes would lower it by collapsing all magnitudes to one value.
Therefore, for $\ell_\perp$ we do not backpropagate through the magnitudes $\r_i^v$. We also treat the bandwidths $\s_\r$ and $\s_{\hat u}$ as constants, so $\ell_\perp$ moves only the directions, through the kernel values $L_{ij}^v$.
We set $\l_\perp$, so that the gradient of $\l_\perp \ell_\perp$ with respect to the embeddings has a similar norm as the gradient of $\l_{\hat u} \ell_{\hat u}$.

In \cref{tab:independence}, we compute \cref{eq:independence} on one view of 3000 validation images, with bandwidths $\s_\r$ and $\s_{\hat u}$ from a random subset of 2000 of them.
Raw values of $\text{MMD}^2$ are not comparable across methods, since the bandwidths are set from the embeddings of each method.
We divide the result by its standard deviation over 200 shuffles of the magnitudes across images, which make magnitude and direction independent; this ratio is comparable across methods.

\subsection{The coefficients of PolarReg-naive can be tuned once and scaled to new settings}
\label{app:autodose}

The angular coefficient $\l_{\hat u}$ and magnitude coefficient $\l_\r$ of PolarReg-naive have a broad optimum.
We tune the coefficients once, on Tiny-ImageNet-200 with $\bb = 256$ and $d = 64$, to $\l_{\hat u} = 20$ and $\l_\r = 0.045$.
In \cref{tab:app:weights}, scaling tuned coefficients by 0.25--4$\times$, the resulting linear probe changes by at most 1 point in accuracy.

The coefficients can be rescaled to a new setting or dataset with a gradient measurement.
The coefficients $\l_{\hat u}$ and $\l_\r$ in \cref{eq:full_decomposition} weigh the regularizers with respect to the alignment term. This conversion changes with the batch size, embedding dimension and data.
For example, for $n$ independent samples of a distribution with CDF $F$ and density $f$, the expected $W_2^2$ to that distribution falls as $1/n$ if $\int F(1-F)/f < \infty$~\cite[Theorems 5.1 and 5.9]{bobkov2019one}, which holds for the inner products. For the angular regularizer, samples are the $\bb(\bb-1)/2$ pairs. A coefficient tuned for such a regularizer at $\bb = 256$ acts about 15 times stronger at $\bb = 64$.
In a new setting, we measure the gradient norms $g_{\text{align}}$, $g_{\hat u}$ and $g_\r$ of the three terms at the first step after warm-up. We rescale the coefficients $\l_{\hat u}$ and $\l_\r$ so that the terms keep the gradient ratios they had in the tuned setting,
\beq{
    \l_{\hat u}' = \l_{\hat u}\, \frac{g_{\hat u}\, g_{\text{align}}'}{g_{\text{align}}\, g_{\hat u}'},
    \qquad
    \l_\r' = \l_\r\, \frac{g_\r\, g_{\text{align}}'}{g_{\text{align}}\, g_\r'}.
    \label{eq:autodose}
}
where $\l'$ and $g'$ are the coefficients and gradient norms in the new setting. The rescaled coefficients are then kept fixed.
On CIFAR-10, rescaling by \cref{eq:autodose} recovers the coefficients within 4\% of the tuned coefficients.
At ImageNet-1k, it is very expensive in terms of compute, to search for the optimum coefficients. In all our PolarReg-naive, ImageNet-1k experiments, we report performance with the coefficients rescaled from Tiny-ImageNet-200.

\begin{table}[h]
    \centering
    \caption{\textbf{Both coefficients work over a broad range around their tuned values.}
    Linear-probe accuracy (\%) at the last epoch when one coefficient is scaled and the other is kept at its tuned value, $\l_{\hat u} = 20$ or $\l_\r = 0.045$.
    Tiny-ImageNet-200, $\bb = 256$, $d = 64$, 200 epochs, one seed per cell.}
    \label{tab:app:weights}
    \begin{tabular}{lccccc}
        \toprule
        Scale of the coefficient & 1/4 & 1/2 & 1 & 2 & 4 \\
        \midrule
        Angular coefficient $\l_{\hat u}$ & 52.51 & 52.53 & 52.41 & 52.42 & 51.33 \\
        Magnitude coefficient $\l_\r$ & 52.27 & 53.23 & 52.41 & 52.87 & 52.44 \\
        \bottomrule
    \end{tabular}
\end{table}

\subsection{Clipping each optimizer step is a simple way to stabilize training of PolarReg-naive and sliced objectives}
\label{app:trust_region}

In some PolarReg-naive and SIGReg runs, training collapses. A collapse starts with an update of ordinary gradient norm that moves the embeddings far from their target distribution, and the regularizer amplifies the change in a few steps. Gradient clipping is a poor fix. Its threshold depends on the scale of the loss, which differs 30 times across our datasets.
A threshold set too low  or high, slows training or collapses it.

Adam divides each coordinate of the gradient by a running estimate of its scale, making the step approximately scale independent.
The running estimate averages over about 1000 steps, so a sudden change, such as the start of a collapse, makes the step unusually large.
We therefore clip the Adam step instead of the gradient, like Adafactor~\cite{shazeer2018adafactor}. We rescale the step $\D\th$ whenever it exceeds $\tau = 2$ times the median $\mu_t$ of recent step sizes \cite{seetharaman2020autoclip},
\beq{
    \D\th \leftarrow \D\th \cdot \min\rbr{1, \frac{\tau\, \mu_t}{\norm{\D\th}}}.
    \label{eq:trust_region}
}
For our experiments, in standard training, the 99th percentile of $\norm{\D\th}$ is 1.3--1.5 times its median, against 4.5--33 times for the gradient norm. During a collapse, for PolarReg-naive or SIGReg, the step size $\norm{\D\th}$ reaches 12--20 times its median.
Setting $\tau = 2$ prevented every collapse we observed. The threshold needs no tuning. It prevents collapse on both Tiny-ImageNet-200 and the ImageNet-100, whose tuned gradient clips differ by a factor of 10.

\textbf{We note that PolarReg requires no such optimizer step bounds for stable training.}

\subsection{Training details}
\label{app:training}

\paragraph{ImageNet-1k pretraining.}
We pretrain ViT-B/16 and ViT-L/14 encoders with the recipe of VISReg~\cite{wu2026visreg}. We do not do any hyperparameter search.
Each batch holds 512 images, and we make 4 global views at 224 px and 6 local views at 96 px for each image. Every view is a random resized crop, with scale $[0.3, 1]$ for global views and $[0.05, 0.3]$ for local views. It is then color jittered with probability 0.8 (brightness 0.4, contrast 0.4, saturation 0.2, hue 0.1), converted to grayscale with probability 0.2, blurred with a Gaussian of standard deviation in $[0.1, 2]$ px with probability 0.5, solarized with probability 0.2, and flipped horizontally with probability 0.5.
The encoder output is mapped to the embeddings $u_i^v$ of \cref{eq:lejepa} by a projector, an MLP with two hidden layers of width 2048, batch normalization and GELU activations. It takes the concatenated CLS tokens of the last two blocks, and outputs $d = 256$ dimensions for ViT-B and $d = 384$ for ViT-L.
We use the AdamW optimizer at a peak learning rate of $9 \times 10^{-4}$ for ViT-B/16 and $8 \times 10^{-4}$ for ViT-L/14, with a linear warm-up over 5 epochs followed by a cosine decay to $1/1000$ of the peak learning rate.
PolarReg uses $\eta_{\hat u} = 2$ and $\eta_\r = 3$.
The linear probes on ImageNet-1k follow VISReg~\cite{wu2026visreg}. Classifiers on the concatenated CLS tokens of last four blocks of the frozen encoder are trained for 100 epochs with batch size 256. We report accuracy on the 50,000 validation images.
In \cref{tab:in1k}, all rows except ours are from VISReg~\cite[Table 5]{wu2026visreg}.

\paragraph{Compute.}
For ImageNet-1k training, we use NVIDIA B200 GPUs to train our models. Pretraining for 400 epochs consumes 135 B200 GPU hours for ViT-B and 559 for ViT-L. VISReg reports about 1120 and 2060 H100 GPU hours for the same models and epochs. We use NVIDIA DALI for efficient data loading; bfloat16 and \texttt{torch.compile} for faster training.

\paragraph{Transfer to other datasets.}
For other datasets, we follow VISReg~\cite{wu2026visreg}.
The features are the concatenated CLS tokens of the last four blocks of the frozen encoder, computed on images resized to $224 \times 224$ pixels without augmentation.
They are standardized with batch statistics (BatchNorm without learnable parameters). A linear classifier trains on them for 10 epochs with batch size 32 at 13 learning rates.
For ChestX-ray14, where an image can have several of the 14 findings, the classifier is trained with a binary cross-entropy loss, and we report the mean average precision at the learning rate with the best mean AUROC.
Aircraft, Stanford Cars, CIFAR-10, CIFAR-100, Flowers-102, Food-101, Oxford-IIIT Pets and DTD use their standard train and test splits. ChestX-ray14, and RetinaMNIST and OrganAMNIST from MedMNIST+ at 224 pixels, use their official splits.

\paragraph{ImageNet-100 pretraining.}
For \cref{tab:independence}, we train ViT-B/16 encoders on ImageNet-100~\cite{tian2020cmc} for 100 epochs, each on one RTX 4090.
Each batch holds 256 images, and each image creates 2 global views at 224 px and 6 local views at 96 px. The projector outputs $d = 256$ dimensional embeddings.
We train with AdamW at a learning rate of $5 \times 10^{-4}$, with one warm-up epoch and a cosine decay to 0.
SIGReg uses $\lambda = 0.05$ and 4096 slices; PolarReg-naive uses $\lambda_{\hat u} = 14.3$ and $\lambda_\r = 0.062$.
The augmentations differ from those on ImageNet-1k in three numbers only: global crop scale $[0.08, 1]$, local crop scale $[0.05, 0.4]$, and color jitter strengths $(0.8, 0.8, 0.8, 0.2)$.
The probe follows the same recipe as the ImageNet-1k probe, but trained for 10 epochs.

\paragraph{Galaxy10 pretraining.}
For \cref{tab:app:galaxy}, we train on Galaxy10 DECaLS. It contains images of galaxies in 10 classes, at 256 px. We use the train and test splits of its Hugging Face release\footnote{\url{https://huggingface.co/datasets/matthieulel/galaxy10_decals}}, with 15,962 and 1774 images.
We train ViT-S/8 encoders from scratch at 128 pixels, with the augmentations of ImageNet-100.
We train with a batch size of 256 images. Each image is used to produce 4 global views and no local views. The projector outputs $d = 256$ dimensional embeddings.
We pretrain the encoders with AdamW at a learning rate of $10^{-3}$, with 10 warm-up epochs and a cosine decay to 0. As VISReg does, we report the best test accuracy of the online linear probe during training.

\section{Supplementary results}
\label{app:supp}

\subsection{The embedding magnitude can be used to detect unusual images}
\label{app:radius}

The embedding magnitude provides a test for images not resembling the training data. This test does not need any calibration data.
For each image, we test the null hypothesis that its magnitude is a draw from $\chi_d$, with a one-sided test on the lower tail. The test rejects the null hypothesis, when its p-value is below $\alpha$.
We choose only the lower tail because we expect unusual images to have larger view disagreement in their embeddings, and such images will attain smaller magnitudes (\cref{prop:rearrangement}).

We apply this test to the ViT-B encoder trained with PolarReg on ImageNet-1k for 400 epochs, with $d = 256$.
We term the 50,000 ImageNet-1k validation images as ``in-domain''. The ``out-of-domain'' images are 5000 of ImageNet-1k validation images with different levels of corruptions, and images from datasets of other domains.
For each out-of-domain set, we report the area under the curve of the fraction of out-of-domain images the test rejects against the fraction of in-domain images it rejects, as $\alpha$ goes from 0 to 1. This area is simply the AUROC; 0.5 means the test detects out-of-domain images no better than chance, and 1 means some suitable $\a$ detects all of them without flagging any in-domain image.

For corruptions in \cref{tab:app:radius}, blur convolves the image with a Gaussian of standard deviation $\sigma$ pixels. A contrast change maps each pixel value $x \in [0, 1]$ to $0.5 + c\,(x - 0.5)$, for some contrast $c$. Block averaging replaces each $f \times f$ block of pixels with its mean.
The test detects significant corruptions to the ImageNet-1k validation images.
It detects images averaged over $8 \times 8$ blocks with an AUROC of 0.99.
Minor corruptions are not detected.
Blur with $\sigma = 1$ pixel and halved contrast are detected with an AUROC of 0.50.
This is perhaps because the encoder is trained to be invariant to such augmentations.
Images from datasets of other domains are detected with high AUROC.

\begin{table}[h]
    \centering
    \caption{\textbf{The lower-tail test of the embedding magnitude distinguishes corrupted images and out-of-domain data from in-domain data.}
    AUROC of one-sided test of the embedding magnitude against $\chi_{256}$, with ImageNet-1k validation images as in-domain images, for a ViT-B/16 PolarReg encoder trained for 400 epochs with $d = 256$.}
    \label{tab:app:radius}
    \small
    \begin{tabular}{lc}
        \toprule
        Out-of-domain dataset & AUROC \\
        \midrule
        \textbf{ImageNet-1k validation images, corrupted by} & \\
        \quad Blur, $\sigma = 1$ & 0.50 \\
        \quad Blur, $\sigma = 4$ & 0.76 \\
        \quad Contrast, $c = 0.5$ & 0.50 \\
        \quad Contrast, $c = 0.1$ & 0.76 \\
        \quad Block averaging, $f = 4$ & 0.90 \\
        \quad Block averaging, $f = 8$ & 0.99 \\
        \midrule
        \textbf{Datasets from other domains} & \\
        \quad Galaxy10 (galaxies) & 0.80 \\
        \quad AID (aerial scenes) & 0.81 \\
        \quad ChestX-ray14 (chest X-rays) & 0.82 \\
        \quad RetinaMNIST (retina images) & 0.95 \\
        \quad OrganAMNIST (abdominal CT) & 0.91 \\
        \bottomrule
    \end{tabular}
\end{table}

\subsection{The angular regularizer $\ell_{\hat u}$, in \cref{eq:angular}, detects duplicated images in a batch}
\label{app:duplicates}
Large image datasets, whether crawled from the web or recorded in the real world for robotics, contain many near-duplicate images. For instance, DINOv2 \cite{oquab2024dinov2} removed about 200 million of its 1.3 billion crawled images as near-duplicates.
We test whether the angular regularizer and SIGReg flag duplicates, on synthetic batches and on trained encoders.
We compute each criterion on 1000 control batches of 256 samples of $\text{N}(0, I_{d \times d})$, and use the 95th percentile of these values as its threshold. A duplicated batch is a control batch in which the directions of $p$ samples are replaced by the directions of $p$ others; $p$ pairs share a direction. We repeat this 5 times with fresh samples and report how many times the criterion exceeds its threshold.
Even with $p = 1$, the angular regularizer is 6.2 times its mean on control batches at $d = 64$, and exceeds its threshold in all 5 repeats.
SIGReg with 4096 slices stays within 8\% of its mean up to 13 pairs, and exceeds its threshold in all constructions only at 32 pairs, a quarter of the batch.
Its values follow $1 + 2p/\bb$ expectation derived in \cref{rem:app:duplicates} within 6\%.

For trained encoders, we duplicate an image by adding a second augmented view of it to the batch.
On PolarReg-naive trained on Tiny-ImageNet-200 with $d = 64$, duplicating 5\% of each batch raises the angular regularizer above three standard deviations of its value on clean batches when averaged over 7 such duplicated batches. For a similar deviation with 10\% duplication, it only needs one duplicated batch.
SIGReg needs 22 batches to reach three standard deviations at 10\%, and never reaches it at 5\% or below.

\subsection{PolarReg outperforms PolarReg-naive, without any tuning}
\label{app:eta}
PolarReg-naive needs its coefficients $\l_{\hat u}$ and $\l_\r$ tuned for each setting, while PolarReg uses the same step sizes $\eta_{\hat u} = 2$ and $\eta_\r = 3$ everywhere.
As described in \cref{app:autodose}, we tuned the PolarReg-naive coefficients on Tiny-ImageNet-200 with $\bb = 256$ and $d = 64$, and rescaled them to other settings. In \cref{tab:app:eta}, PolarReg outperforms PolarReg-naive in every setting.

\begin{table}[h]
    \centering
    \caption{\textbf{PolarReg outperforms PolarReg-naive in every setting without any changes to $\eta_{\hat u}$ or $\eta_\r$.}
    Linear-probe accuracy (\%) at the final epoch. The coefficients of PolarReg-naive were tuned in the setting of the first row and rescaled for other rows.
    }
    \label{tab:app:eta}
    \begin{tabular}{lccc}
        \toprule
        Setting & Epochs & PolarReg-naive & PolarReg \\
        \midrule
        Tiny-ImageNet-200, $\bb = 256$, $d = 64$ & 400 & 53.46 & 54.55 \\
        Tiny-ImageNet-200, $\bb = 256$, $d = 256$ & 400 & 54.25 & 55.43 \\
        Tiny-ImageNet-200, $\bb = 64$, $d = 64$ & 200 & 32.52 & 36.89 \\
        ImageNet-1k, ViT-B/16 & 100 & 74.50 & 75.50 \\
        \bottomrule
    \end{tabular}
\end{table}

PolarReg only needs $\eta_{\hat u}$ to be large enough; $\eta_{\hat u}$ sets how far along the target direction lies.
With a very low $\eta_{\hat u} = 0.5$, the target is too close to pull the directions apart, against the alignment term and training collapses.
Above that, accuracy is stable.
We choose $\eta_\r = 3$ to calibrate the mean magnitude close to $\sqrt d$, which is about the mean of $\chi_d$.

\paragraph{PolarReg is robust across step sizes.}
On ImageNet-100, we train a ViT-B/16 for 100 epochs with $d = 256$, and change one step size at a time while the other stays at its default, $\eta_{\hat u} = 2$ or $\eta_\r = 3$.
We try $\eta_{\hat u} = $ 1, 1.5 and 3, and $\eta_\r = 2$ and 4.
In \cref{tab:app:eta_sweep}, accuracy changes by at most 0.2 points when $\eta_\r$ changes, and by at most 1.6 points when $\eta_{\hat u}$ changes.
On ImageNet-1k, we train a ViT-B/16 for 400 epochs with both step sizes 1.5 times their defaults, $\eta_{\hat u} = 3$ and $\eta_\r = 4.5$.
Its linear-probe accuracy is 76.55\%, against 76.65\% with the defaults.

\begin{table}[h]
    \centering
    \caption{\textbf{PolarReg is robust across step sizes.}
    Linear-probe accuracy (\%) at the last epoch when one step size is changed and the other is kept at its default, $\eta_{\hat u} = 2$ or $\eta_\r = 3$.
    }
    \label{tab:app:eta_sweep}
    \begin{tabular}{lcccc}
        \toprule
        Angular step size $\eta_{\hat u}$ & 1 & 1.5 & 2 & 3 \\
        Accuracy (\%) & 82.56 & 81.64 & 82.42 & 80.84 \\
        \midrule
        Magnitude step size $\eta_\r$ & 2 & 3 & 4 & --\\
        Accuracy (\%) & 82.26 & 82.42 & 82.44 & -- \\
        \bottomrule
    \end{tabular}
\end{table}

\subsection{PolarReg training generalizes to other modalities}
\label{app:multimodal}

\begin{figure}[t]   %
    \centering
    \captionsetup[subfigure]{labelfont=bf}
    \begin{subfigure}{\linewidth}
        \centering
        \includegraphics[width=\linewidth]{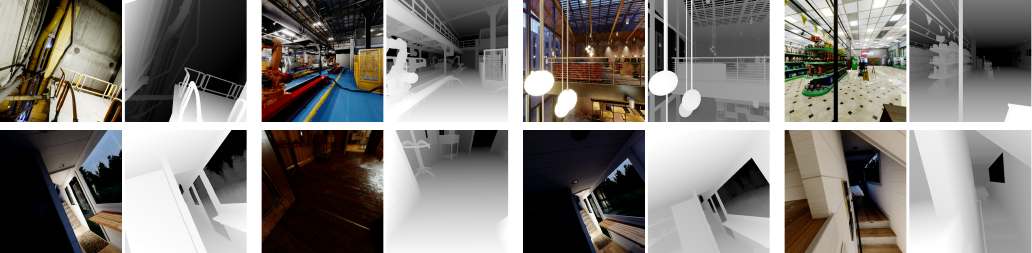}
        \caption{Example RGB and depth images from TartanAir. Top row and bottom row show daytime and nighttime samples. Depth is gray, with closer objects brighter.}
        \label{fig:app:multimodal:examples}
    \end{subfigure}
    \vspace{6pt}
    \begin{subfigure}{\linewidth}
        \centering
        \small
        \begin{tabular}{lcc}
            \toprule
            & \multicolumn{2}{c}{Accuracy (\%)} \\
            \cmidrule(lr){2-3}
            Method & Daytime & Nighttime \\
            \midrule
            RGB encoder, RGB classifier & 95.9 & 30.3 \\
            Depth encoder, depth classifier & 79.3 & 71.6 \\
            Depth encoder, RGB classifier, no map & 3.5 & 7.6 \\
            Depth encoder, RGB classifier, orthogonal map & 75.1 & 68.5 \\
            \bottomrule
        \end{tabular}
        \caption{Scene classification accuracy of a linear classifier trained on frozen encoders of daytime RGB and depth images on TartanAir; tested on daytime and nighttime data.}
        \label{fig:app:multimodal:table}
    \end{subfigure}
    \caption{\textbf{PolarReg featurizes depth images with the same objective as RGB images.}
    (a) Example RGB and depth images from TartanAir. (b) Scene classification accuracy of linear classifiers on frozen features.
    Depth trails RGB on daytime data, but stays accurate at night, when RGB fails.
    An orthogonal map, fitted on unlabeled RGB and depth image pairs, aligns depth features with RGB features. Then, the RGB classifier is almost as accurate on depth images as the depth classifier, without any ground truth labels for depth.}
    \label{fig:app:multimodal}
\end{figure}

To show that PolarReg can be applied to other data modalities, we train an encoder on depth images with the same objective and hyperparameters as on RGB images.
We use the RGB and depth images of the easy trajectories (``Data\_easy'') of TartanAir v2~\cite{patel2025tartanground}, in 18 scenes, at $64 \times 64$ pixels.
We train on 126 trajectories and test on 30.
We never train and test on the same trajectory, repeated at different times of day and seasons, since these render the sensors at same viewpoints.
Five of these 18 scenes also have nighttime data. We train on the 38,487 daytime frames, and test on 5533 daytime and 1319 nighttime frames. \Cref{fig:app:multimodal:examples} shows some examples.
We train one ViT-S/8 encoder per modality from scratch for 100 epochs with PolarReg, $\eta_{\hat u} = 2$ and $\eta_\r = 3$, as in every other experiment.
We take logarithm of distance as input for the depth encoder.
A linear classifier on frozen features predicts the scene where the RGB or depth image is captured from. Therefore, there are 18 classes to predict from.

\paragraph{PolarReg trains a useful depth encoder with no change to the objective (\cref{fig:app:multimodal}b).}
Its classifier is 79.3\% accurate on daytime frames, against 95.9\% for RGB.
At night, the RGB classifier, trained only on daytime images, falls to 30.3\%, while depth stays at 71.6\%. Of course, this is expected given the depth sensor is invariant to lighting in TartanAir.

\paragraph{A classifier trained for one modality can work quite well for another, after an orthogonal map between their frozen features.}
We fit an orthogonal map from depth features to RGB features, trained independently as described above.
We use only 8192 paired daytime frames to learn this map on the frozen features, standardized along each feature dimension.
The frozen RGB classifier is applied to the transformed depth features, achieving an accuracy of 75.1\% on daytime and 68.5\% on nighttime depth images.
Since the RGB and depth encoders are trained separately, nothing aligns their features, and transfer fails without the map.
So even without any ground truth labels for one modality, we can classify its data with a classifier trained for another modality, using only a few unlabeled pairs.

A full multimodal training needs little beyond PolarReg.
With many paired sensor data, one could apply PolarReg to the joint features of both modalities directly, essentially treating them as a single modality.
We leave this, and sensors such as LiDAR, event cameras and radar, to future work.

\subsection{PolarReg generalizes across domains, tasks and dataset scales}
\label{app:beyondin1k}

\paragraph{PolarReg frozen features transfer to out-of-domain datasets and segmentation tasks.}
We probe frozen features of encoders pretrained on ImageNet-1k on six datasets from other domains, and on semantic segmentation of ADE20K. The results are presented in \cref{tab:app:ood}.
The out-of-domain datasets contain textures (DTD), galaxies (Galaxy10), aerial scenes (AID), chest X-rays (ChestX-ray14), retina images (RetinaMNIST) and abdominal CT (OrganAMNIST).
The classifiers are trained as in \cref{app:training}. For ChestX-ray14, where an image can have several of the 14 findings, we report mean average precision (mAP).
The segmentation head is a linear layer on frozen patch features.
For these tasks, PolarReg features perform similar to most other SSL methods, including the sliced objectives.

\begin{table}[h]
    \centering
    \caption{\textbf{Features learned by PolarReg transfer to out-of-domain datasets and segmentation.}
    Probing accuracy (\%) on six out-of-domain datasets, and mean intersection over union (mIoU, \%) of a linear segmentation head on ADE20K, from frozen features of encoders pretrained on ImageNet-1k.
    Rows except ours are reported by VISReg \cite[Tables 6 and 8]{wu2026visreg}.}
    \label{tab:app:ood}
    \small
    \begin{adjustbox}{max width=\textwidth}
    \begin{tabular}{llc|cccccc|c}
        \toprule
        Method & Backbone & Epochs & DTD & Galaxy10 & AID & ChestX-ray & RetinaMNIST & OrganAMNIST & ADE20K \\
        \midrule
        MoCo v3~\cite{chen2021mocov3} & ViT-B/16 & 300 & 73.72 & 73.06 & 90.20 & 23.89 & 64.50 & 91.37 & 31.69 \\
        DINO~\cite{caron2021dino} & ViT-B/16 & 400 & 74.26 & 72.77 & 91.52 & 24.63 & 62.50 & 91.70 & 29.40 \\
        iBOT~\cite{zhou2022ibot} & ViT-B/16 & 400 & 74.10 & 71.65 & 91.73 & 24.80 & 63.75 & 91.84 & -- \\
        I-JEPA~\cite{assran2023ijepa} & ViT-H/14 & 300 & 69.89 & 71.31 & 88.68 & 23.46 & 65.75 & 92.19 & -- \\
        \midrule
        MAE~\cite{he2022mae} & ViT-L/16 & 1600 & 72.77 & 71.98 & 86.42 & 22.87 & 63.00 & 90.06 & -- \\
        LeJEPA (ours) & ViT-B/16 & 100 & 73.56 & 72.04 & 90.47 & 21.41 & 63.50 & 95.82 & 28.94 \\
        VISReg & ViT-B/16 & 400 & 75.69 & 74.01 & 90.91 & 24.88 & 62.25 & 93.40 & 30.16 \\
        VISReg & ViT-L/14 & 400 & 76.54 & 76.32 & 90.12 & 23.64 & 64.25 & 92.93 & -- \\
        \midrule
        PolarReg-naive & ViT-B/16 & 100 & 74.20 & 73.28 & 90.33 & 21.33 & 63.75 & 95.56 & 29.58 \\
        PolarReg & ViT-B/16 & 100 & 75.00 & 72.04 & 90.59 & 22.34 & 61.50 & 95.76 & 29.80 \\
        PolarReg & ViT-B/16 & 400 & 75.96 & 73.45 & 90.42 & 22.21 & 60.50 & 95.46 & 30.50 \\
        PolarReg & ViT-L/14 & 100 & 74.26 & 75.99 & 89.11 & 21.82 & 63.25 & 95.16 & 30.19 \\
        PolarReg & ViT-L/14 & 400 & 76.56 & 76.03 & 89.67 & 23.08 & 64.25 & 95.37 & 30.80 \\
        \bottomrule
    \end{tabular}
    \end{adjustbox}
\end{table}

\paragraph{PolarReg pretrains small models on small datasets better than other methods.}
We pretrain ViT-S/8 encoders from scratch on Galaxy10, images of galaxies in 10 classes, with the recipe of VISReg~\cite{wu2026visreg}. Training details are provided in \cref{app:training}. In \cref{tab:app:galaxy}, with half the epochs, PolarReg matches the performance of other methods. With a full training run of 400 epochs, it outperforms all methods.

\begin{table}[h]
    \centering
    \caption{\textbf{Linear probing accuracy of ViT-S/8 encoders pretrained from scratch on Galaxy10.}
    All encoders are trained for 400 epochs at 128~px, unless marked.
    Rows except PolarReg are from VISReg~\cite[Table 2]{wu2026visreg}.}
    \label{tab:app:galaxy}
    \small
    \begin{tabular}{lc}
        \toprule
        Method & Accuracy (\%) \\
        \midrule
        DINO~\cite{caron2021dino} & 73.49 \\
        VICReg~\cite{bardes2022vicreg} & 79.93 \\
        SIGReg~\cite{lejepa2025} & 80.50 \\
        SWD~\cite{bonneel2015sliced} & 80.60 \\
        VISReg~\cite{wu2026visreg} & 80.51 \\
        \midrule
        PolarReg & 82.13 \\
        PolarReg (200 epochs) & 80.38 \\
        \bottomrule
    \end{tabular}
\end{table}

\subsection{PolarReg is more memory and compute efficient than sliced objectives}
\label{app:cost}
In \cref{tab:app:cost}, we time one forward and backward pass of each objective, on Gaussian embeddings of $\bb$ images with 10 views each and $d = 256$, as in our ImageNet-1k runs.
SIGReg and VISReg use 4096 slices. We report the median of 10,000 calls.
PolarReg is overhead-bound. It is bound by GPU kernel launch overhead, not by the size of the intermediate tensors.
The sliced objectives are memory-bound. 
They are bound by large tensor read-write operations that grow linearly with $\bb$.
They project every embedding on 4096 slices. SIGReg also approximates the integral in \cref{eq:lejepa} by evaluating the empirical characteristic function of each projection at several values of $t$.
Reading and writing these numbers takes time proportional to $\bb$.
PolarReg works with the $\bb\vv$ magnitudes and the inner products between the $\bb$ mean directions.
At these batch sizes, each of its kernels runs faster than PyTorch can launch it, so its time does not grow with $\bb$.
Its backward pass is one elementwise operation, since no gradient flows through the targets.

\begin{table}[h]
    \centering
    \caption{\textbf{Computation time of the regularization objectives on a single RTX 4090.}}
    \label{tab:app:cost}
    \begin{tabular}{lcccccc}
        \toprule
        & \multicolumn{3}{c}{Forward pass (ms)} & \multicolumn{3}{c}{Backward pass (ms)} \\
        \cmidrule(lr){2-4} \cmidrule(lr){5-7}
        Method & $\bb = 224$ & $512$ & $1024$ & $\bb = 224$ & $512$ & $1024$ \\
        \midrule
        SIGReg & 4.95 & 11.10 & 22.06 & 13.74 & 31.34 & 62.65 \\
        VISReg & 1.52 & 3.11 & 6.02 & 0.72 & 1.72 & 3.36 \\
        PolarReg & 0.43 & 0.44 & 0.44 & 0.10 & 0.08 & 0.08 \\
        \bottomrule
    \end{tabular}
\end{table}
We note that these computation times are for eager PyTorch, which launches each kernel separately.
With \texttt{torch.compile}~\cite{ansel2024pytorch2} based CUDA kernel fusion, computation time for the sliced objectives still grows with $\bb$; they stay memory-bound. PolarReg stays overhead-bound, and is still significantly faster.

\paragraph{Memory overhead of the regularizers.} In the forward pass, the sliced objectives compute $\bb\vv\kk$ projections, and SIGReg evaluates each of them at the values of $t$ used to approximate the integral in \cref{eq:lejepa}. All of these are kept for the backward pass.
PolarReg computes $\bb\vv$ magnitudes and $\sim\bb^2$ inner products, and keeps only the $\bb\vv d$ entries of its targets for the backward pass.
Hence, PolarReg needs much less memory in \cref{fig:memory}.

\end{document}